\documentclass{article} % For LaTeX2e
\usepackage{iclr2027_conference,times}

\usepackage{amsmath,amsfonts,bm}

\def\eqref#1{equation~\ref{#1}}
\def\1{\bm{1}}

\DeclareMathAlphabet{\mathsfit}{\encodingdefault}{\sfdefault}{m}{sl}
\SetMathAlphabet{\mathsfit}{bold}{\encodingdefault}{\sfdefault}{bx}{n}

\usepackage{hyperref}
\usepackage{url}

\newcommand{\methodname}{\textsc{a-b-d}}
\newcommand{\act}{Act·\textsc{onomy}}
\newcommand{\numalltrajectory}{345,667}
\newcommand{\numtrajectory}{27,803}
\newcommand{\numbenchmark}{21}
\newcommand{\numtask}{12}
\newcommand{\nummodel}{80}
\newcommand{\numharness}{50}
\newcommand{\numallfeatureF}{211}
\newcommand{\numallfeatureL}{107}
\newcommand{\numallfeature}{318}
\newcommand{\numfeatureF}{31}
\newcommand{\numfeatureL}{48}
\newcommand{\numfeature}{79}

\usepackage{graphicx}
\usepackage{booktabs}
\usepackage{enumitem}
\usepackage{multirow}
\usepackage{amsmath}
\usepackage{longtable}
\usepackage{xcolor}
\usepackage{colortbl}
\usepackage{wrapfig}
\usepackage{soul}

\definecolor{cPlan}{HTML}{2E6DB4}   % Planfulness
\definecolor{cDelib}{HTML}{169C8F}  % Deliberation
\definecolor{cVerb}{HTML}{E4732B}   % Verbosity
\definecolor{cSoc}{HTML}{B93A8C}    % Sociability
\definecolor{cCur}{HTML}{D8433F}    % Curiosity
\definecolor{cAbs}{HTML}{C9A227}    % Absolutism

\colorlet{cPlanHL}{cPlan!25}
\colorlet{cDelibHL}{cDelib!25}
\colorlet{cVerbHL}{cVerb!25}
\colorlet{cSocHL}{cSoc!25}
\colorlet{cCurHL}{cCur!25}
\colorlet{cAbsHL}{cAbs!25}

\newcommand{\hlPlan}[1]{{\sethlcolor{cPlanHL}\hl{#1}}}

\newcommand{\hlSoc}[1]{{\sethlcolor{cSocHL}\hl{#1}}}
\newcommand{\hlCur}[1]{{\sethlcolor{cCurHL}\hl{#1}}}

\renewcommand{\cite}{\citep}

\title{On the Behavioral Traits of LLM Agents}

\newlength{\logoht}
\newlength{\logohtaff}
\newcommand{\aflogo}[1]{
  \raisebox{-0.35ex}{\includegraphics[height=\logoht]{Figures/logo/#1}}}
\newcommand{\afmark}[1]{\,\aflogo{#1}}          % superscript-style marker after a name
\newcommand{\afitem}[2]{
  \raisebox{-0.4ex}{\includegraphics[height=\logohtaff]{Figures/logo/#1}}~#2}

\newcommand{\unsw}{\afmark{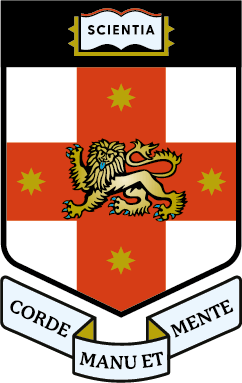}}
\newcommand{\jhu}{\afmark{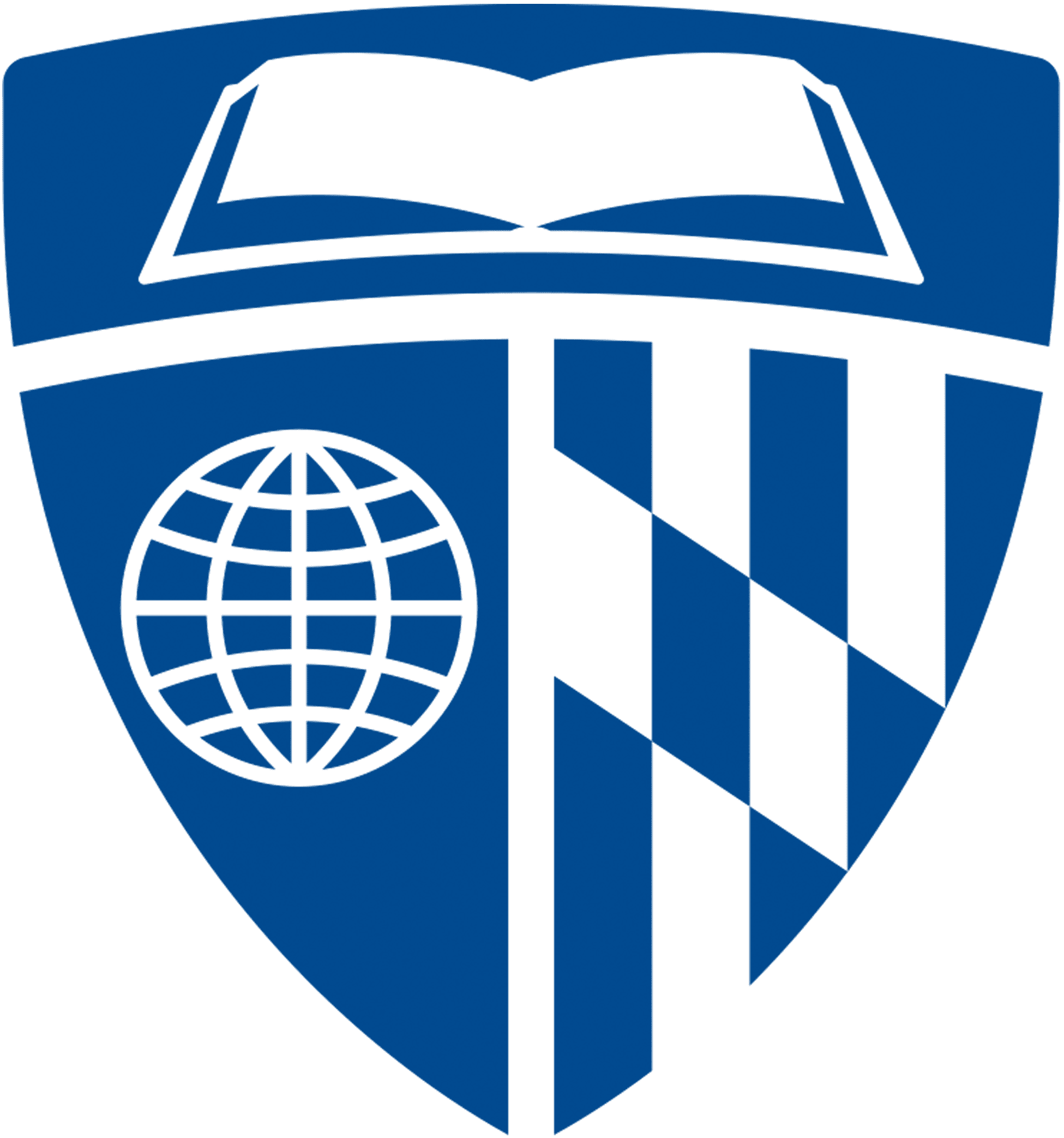}}
\newcommand{\cmu}{\afmark{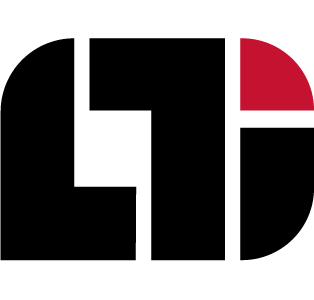}}
\newcommand{\fudan}{\afmark{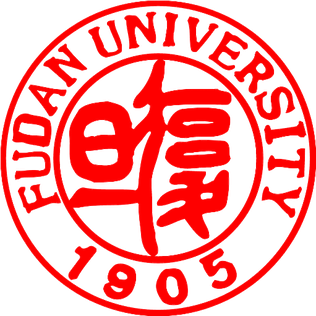}}
\newcommand{\utokyo}{\afmark{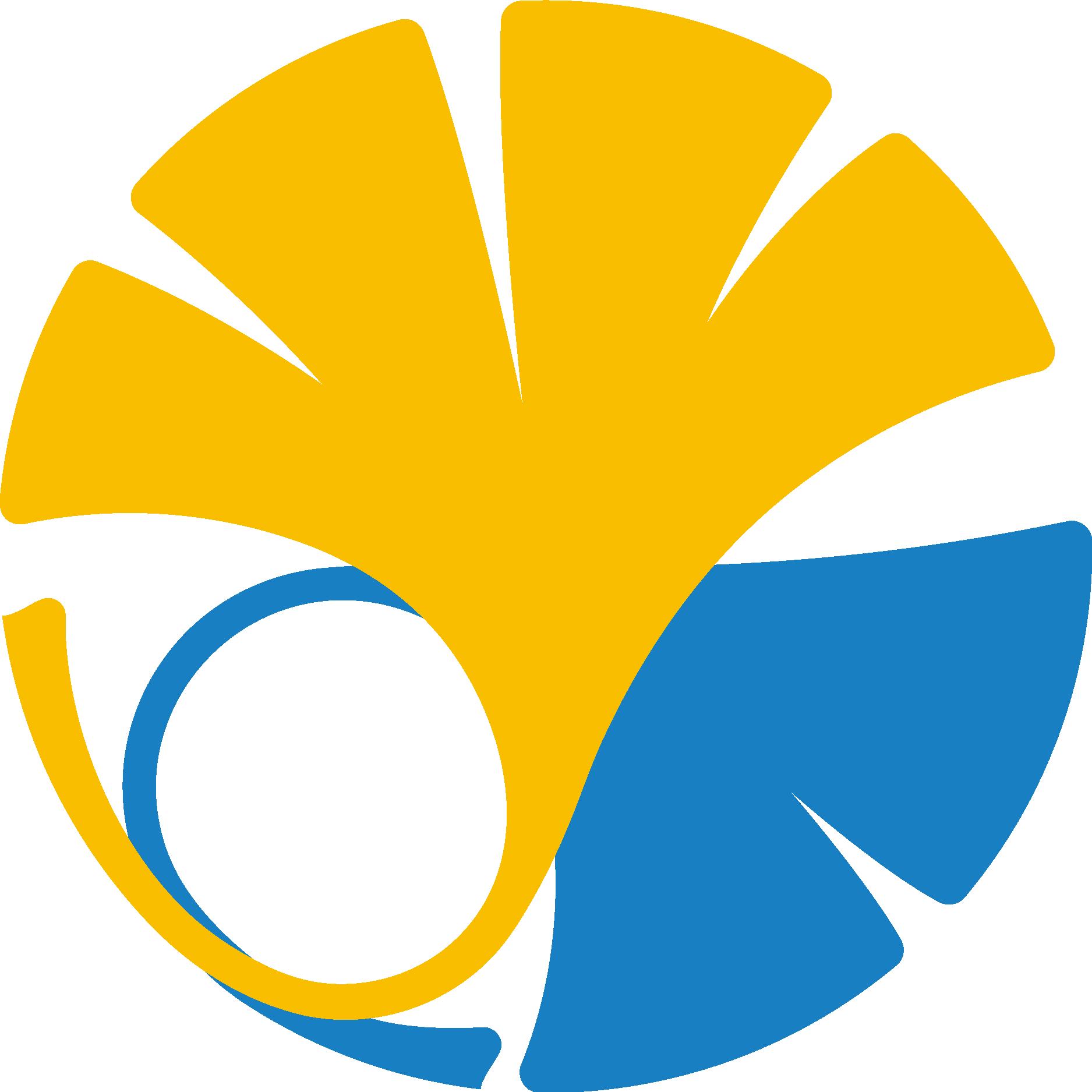}}
\newcommand{\riken}{\afmark{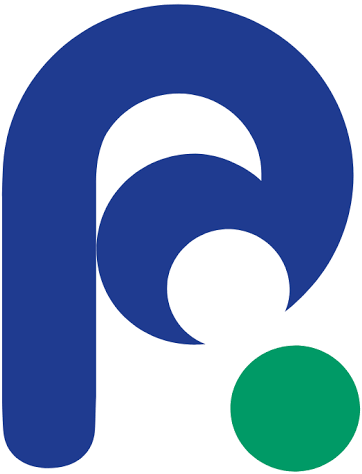}}

\author{
Haokai Zhao\unsw \qquad \quad \; \ Jie Gao\jhu \qquad \quad \; \ Yunze Xiao\cmu \qquad \; Xintao Wang\fudan \\
\bf \ Weihao Xuan\utokyo\riken \qquad Aditya Joshi\unsw \qquad Mark Dredze\jhu \qquad Jen-tse Huang\jhu \\[0.6ex]
{\normalfont\small \afitem{unsw}{University of New South Wales} \quad
 \afitem{jhu}{Johns Hopkins University} \quad
 \afitem{cmu}{Carnegie Mellon University}} \\
{\normalfont\small \afitem{fudan}{Fudan University} \quad
 \afitem{utokyo}{The University of Tokyo} \quad
 \afitem{riken}{RIKEN AIP}} \\[0.6ex]
}

\iclrfinalcopy % Uncomment for camera-ready version, but NOT for submission.

\begin{document}

\maketitle

\vspace{-20pt}
{\footnotesize
\href{https://github.com/JoeZhao527/A-B-D}{\includegraphics[width=10pt]{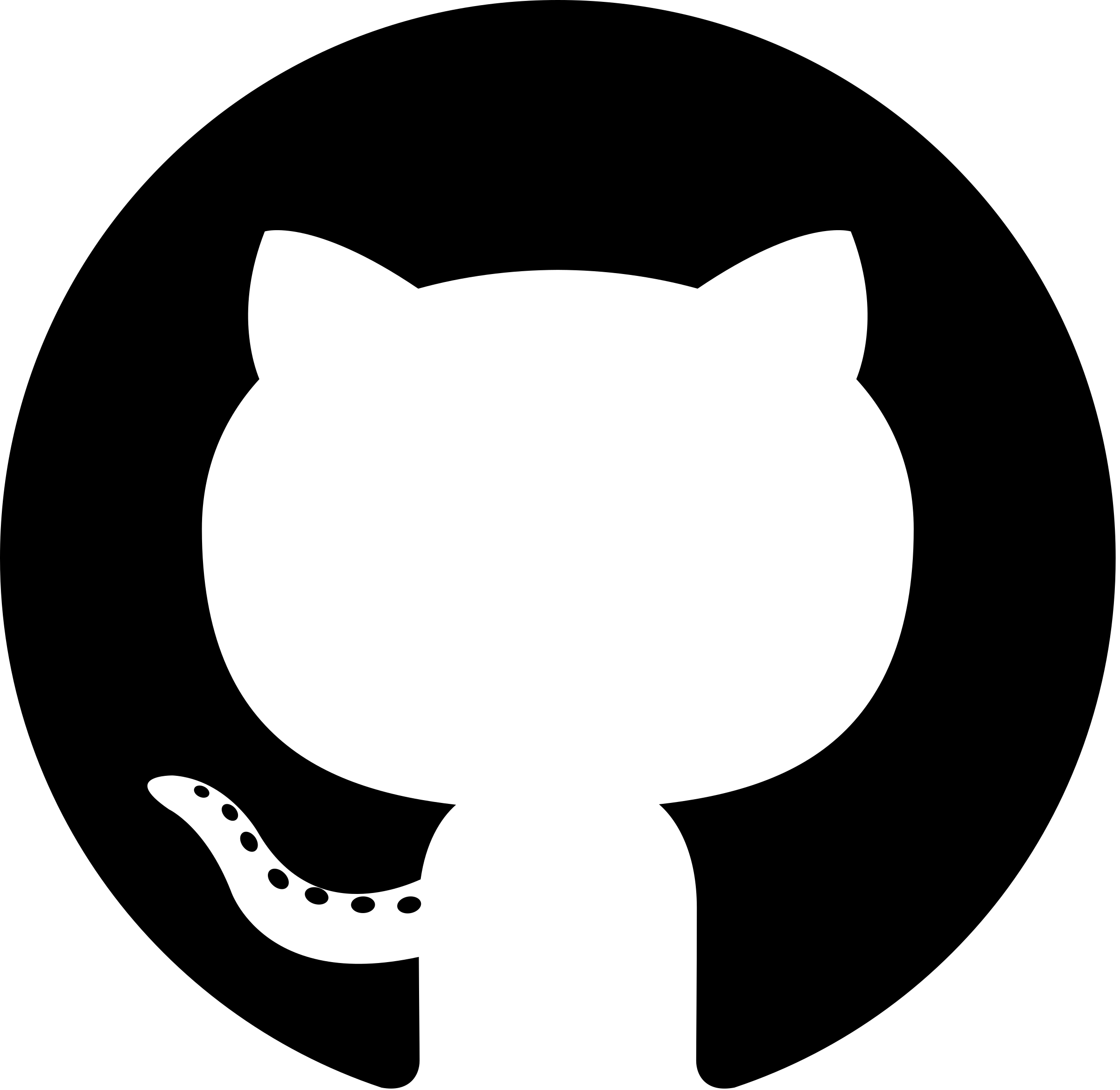}\ \texttt{https://github.com/JoeZhao527/A-B-D}} \ \
\href{https://huggingface.co/A-B-D}{\includegraphics[width=10pt]{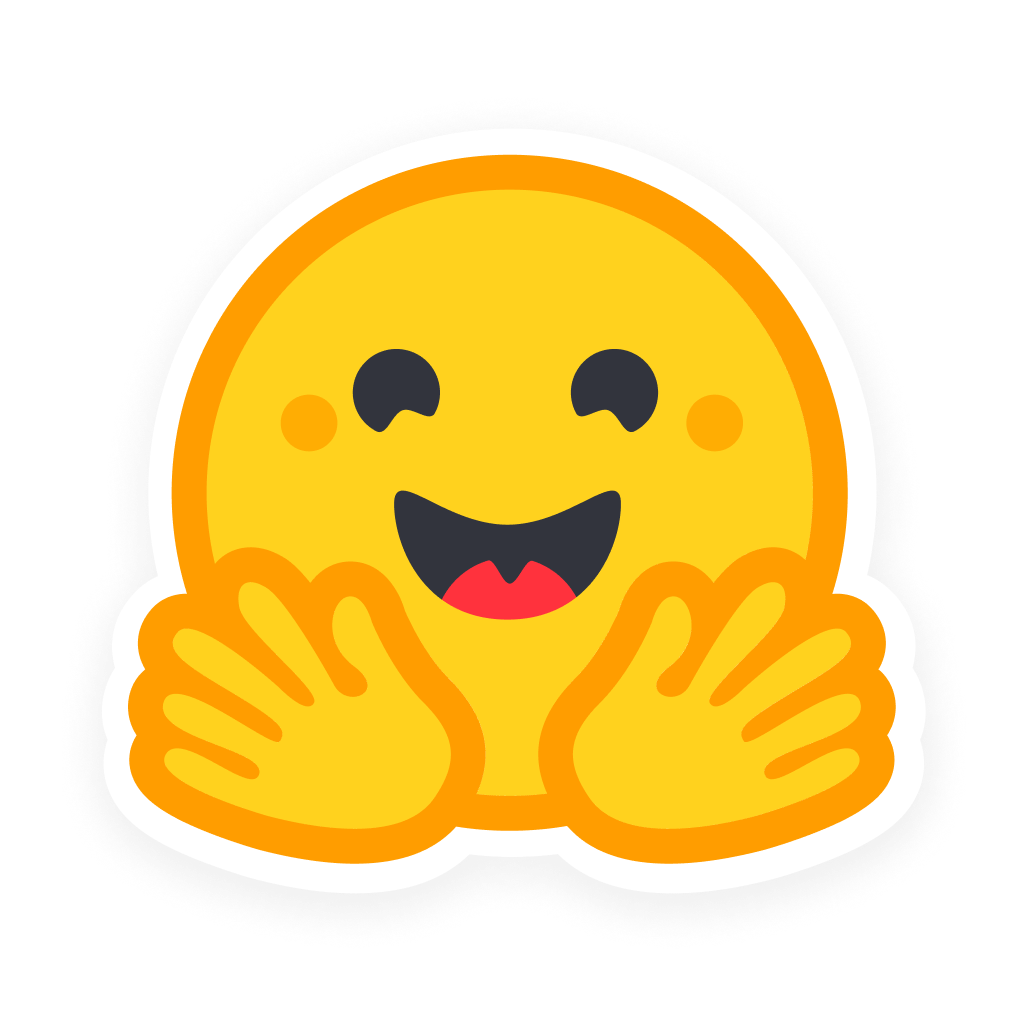}\ \texttt{https://huggingface.co/A-B-D}}
}
\vspace{10pt}

\begin{abstract}

Users increasingly describe different AI agents as distinct colleagues to work with.
AI personality research aims to quantify such impressions by attributing human-like ``traits'' to agents.
However, existing measures fall short: models' self-reports (S-data) diverge from their actual behavior, while informant ratings from LLM judges (I-data) are costly to scale and cover few everyday scenarios.
In this paper, we propose \methodname{} to infer traits bottom-up from behavioral data (B-data), namely how agents act on their environment and communicate with users, as recorded in existing trajectories.
From {\numalltrajectory} real-world trajectories spanning {\nummodel} models, {\numtask} tasks, and {\numharness} harnesses, we extract {\numallfeature} candidate features that capture both the actions an agent takes at each step (functional) and the language accompanying them (linguistic).
We retain only features that show instance-level stability, cross-task consistency, and model discriminability.
Factor analysis of the remaining {\numfeature} features uncovers six stable, model-attributable factors, two functional and four linguistic.
For example, Claude-Sonnet-4.6 and Kimi-K3 exhibit the most planfulness, whereas GPT-5.5 and GPT-5.6 are the least energetic.
Moreover, we quantify the ``knowledge--action gap'' in the wild: these factors correlate only weakly with self-reported Big Five scores, even for conceptually matched pairs such as \textit{extraversion} and \textit{energetic} ($r=0.07, p=0.58$).
Our work offers a new lens for understanding AI personality, with implications for users, developers, and researchers from both computer science and social science.

% (https://arxiv.org/pdf/2501.12326; https://arxiv.org/abs/2507.20534)
\end{abstract}

\section{Introduction}

Users report that AI agents powered by different large language models differ in their initiative, caution, verbosity, and willingness to push back~\cite{reddit2026used, reddit2026does}. These differences in how agents act and communicate give users the impression of distinct ``personalities.'' Such impressions suggest that agent behavior may reflect stable \textit{traits}, tendencies that recur across tasks and distinguish one agent from another. Measuring these traits is useful to agent developers, evaluators, and end users. It can help developers track behavioral changes across model versions, evaluators compare agents beyond task success, and users judge whether an agent's working style fits their needs. We thus ask: \textit{how can we systematically identify and measure these traits?}

AI personality research seeks to measure these traits using methods from personality psychology, which classifies the evidence into four types, collectively termed \textit{BLIS}~\cite{funder2015personality}: self-reports (\textit{S-data}; what the individual says about itself), informant reports (\textit{I-data}; what others say about it), behavioral observations (\textit{B-data}; what it actually does), and life outcomes (\textit{L-data}; the records it accumulates in the world).
Most studies on AI agents rely on S-data, administering questionnaires designed originally for humans, which are easy to operationalize and scale~\cite{huang2024humanity, huang2024reliability, jiang2023evaluating, miotto2022gpt}.
However, recent work reveals a \textit{knowledge--action gap}: the traits LLMs report correlate only weakly with their choices in realistic decision scenarios, and more weakly than in humans~\cite{huang2026knowing, han2025personality, shen2025mind}.
I-data offer a partial solution by having a judge model infer the traits of a target model from their dialogue~\cite{wang2024incharacter, wang2025coser, huang2025beyond}.
Yet they capture only the traits expressed within that dialogue and cannot cover the full range of contexts in which personality manifests.

Questionnaires yield S-data and interviews yield I-data, yet both capture only how a model responds to prompts.
Human B-data, such as naturalistic audio sampling~\cite{mehl2001electronically}, is inherently sparse and subject to observer reactivity~\cite{baumeister2007psychology}.
AI agents, by contrast, produce massive, complete, and replayable trajectories recording how they think, reason, call tools, evaluate their results, and reflect on their outcomes~\cite{liu2026scalecua, wang2025opencua}.
Therefore, these trajectories make B-data available at scale for the first time.

In this work, we introduce {\methodname} (\textsc{A}gent \textsc{B}ehavioral-\textsc{D}ata), a data-driven, bottom-up framework that derives a low-dimensional, model-attributable behavioral structure from in-the-wild agent trajectories using established psychometric methods.
We first collect a large-scale corpus of {\numalltrajectory} trajectories from publicly released runs spanning {\numbenchmark} benchmarks, {\nummodel} models, and {\numharness} harnesses.
Following standard practice in personality psychology, we convert each trajectory into a sequence of predefined behaviors~\cite{goodenough1928measuring, bales1950interaction, buss1983act, funder2000riverside}.
Specifically, we annotate each trajectory using {\act}~\cite{gao2026interpret}, a recent, publicly available, domain-agnostic taxonomy, and generate {\numallfeatureF} functional features.
However, mapping raw outputs to predefined behaviors discards linguistic features, which are known to reflect individual traits~\citep{gottschalk1969measurement, pennebaker1999linguistic}.
We therefore complement the functional features with {\numallfeatureL} linguistic features extracted by LIWC-22~\citep{boyd2022development}.

We retain {\numfeature} features that satisfy three criteria.
(i) \textit{Instance-level stability}: a feature must be expressed consistently across instances within a task rather than driven by particular test cases, which we measure by split-half reliability~\cite{spearman1910correlation}.
(ii) \textit{Cross-task consistency}: a feature must generalize across tasks rather than being task-specific, which we measure by inter-task correlation~\cite{epstein1979stability}.
(iii) \textit{Model discriminability}: a feature must reliably distinguish between models, which we measure by the intra-class correlation coefficient (ICC)~\cite{shrout1979intraclass}.
We then perform factor analysis separately on the functional and linguistic features to summarize them as a few behavioral dimensions.

Our analysis shows that agent behavior in real-world tasks is structured along a small number of stable, model-attributable factors: two functional (planful and deliberate) and four linguistic (talkative, responsible, energetic, and absolutist).
Each model's scores on these factors form its behavioral profile, describing how agents powered by that model tend to act and communicate.
For example, GPT-5.1 is the least planful, Claude-3.5-Sonnet the most responsible, while Gemini-3-Pro and DeepSeek-V3.2 stand out for absolutist.
These profiles are largely invisible to S-data: the factors correlate only weakly with self-reported Big Five scores (mean $|r| = 0.11$; none significant after Bonferroni correction), even for conceptually matched pairs such as extraversion and energetic ($r = 0.07, p = 0.58$).
We further find that behavioral profiles shift across model generations.
Newer models exhibit more planfulness, which also correlates with task success, but score lower on responsible and energetic.
This shift may reflect agentic training that favors explicit planning over the social and affective language characteristic of earlier models.
Our contributions are as follows:
\begin{itemize}[nosep]
\item \textbf{Data.} We construct a large-scale corpus of {\numalltrajectory} real-world agent trajectories spanning diverse models, tasks, and harnesses, curated with quality control and stratified sampling.
\item \textbf{Method.} We introduce {\methodname}, a generalizable and scalable analysis pipeline for characterizing LLM agents through their behaviors. It extracts functional and linguistic features, selects features that are task-stable and model-discriminative, and uses factor analysis separately for each feature family to construct behavioral traits. %Our implementation is available at \url{https://github.com/JoeZhao527/A-B-D}.
\item \textbf{Findings.} We show that, after controlling for task and harness variance, agent behavior exhibits a stable structure attributable to the underlying model. Building on this, we provide, to our knowledge, the first in-the-wild quantification of the ``knowledge--action gap,'' without relying on purpose-built scenarios.
\end{itemize}
\section{The {\methodname} Framework}

% Our goal is to identify and measure stable, model-attributable behavioral traits in agents' actions and language use, based on in-the-wild trajectories. First, we collect and curate trajectories (\S\ref{sec:wild_traj}). Next, we extract functional and linguistic features (\S\ref{sec:feat_construct}). We then adjust for task and harness effects and select features with instance-level stability, cross-task consistency, and model discriminability (\S\ref{sec:stat_tools}). Finally, we use factor analysis to derive behavioral dimensions and score each model on them (\S\ref{sec:factor}).

\subsection{Trajectories in the Wild}
\label{sec:wild_traj}

\paragraph{Collection.}
To identify features that generalize beyond individual tasks, we collect trajectories spanning as diverse a set of scenarios as possible.
We draw on {\numbenchmark} benchmarks that publicly release complete agent trajectories, yielding {\numalltrajectory} trajectories in total.
These sources are heterogeneous, including official leaderboard submissions~\citep{li2026long}, community contributions~\citep{merrill2026terminal}, and independent reproductions~\citep{wang2025openhands}.
This breadth allows us to observe the same model across many tasks and harnesses, and different models on the same task.
We further categorize the benchmarks into {\numtask} tasks, enabling stratified sampling so that no single category dominates the dataset.
Categories include dialogue~\citep{yao2025tau, barres2026tau}, tool use~\citep{li2026tool}, software engineering~\citep{zan2025multi}, and research~\citep{wang2026naturebench}.
Table~\ref{tab:traj_sources_filtered} summarizes the statistics of each benchmark.

\paragraph{Standardization.}
Each source records trajectories in its own format, such as chat message lists, tool-call spans, or reasoning--action iterations.
We first parse each format into a unified event stream, in which each event is a system prompt, a user or agent message, a reasoning trace, a tool call, or an observation, and attach a header containing task and outcome metadata.
We then map the model, harness, and task names used by each source to canonical identifiers, so that an entity recorded under multiple aliases is counted once.
Finally, we serialize every trajectory into the Agent Trajectory Interchange Format (ATIF v1.8)~\citep{harbor2026framework}, in which each step comprises a single model call (its reasoning, message, and tool calls) together with the observations; images are replaced with placeholders.
All trajectories pass the ATIF validator, and all subsequent analyses operate on this standardized data.

\paragraph{Quality control.}
We apply a sequence of filters to remove low-quality trajectories; Appendix~\ref{app:funnel} reports the number of trajectories remaining after each stage.
Specifically, we discard trajectories
(i) produced by models outside our target population, such as task-specific fine-tunes and intermediate checkpoints;
(ii) whose harness cannot be identified, such as those from customized harnesses;
(iii) that contain too little behavior to annotate, including result-only records, fewer than five events or fewer than five steps, and runs without genuine agent actions (no tool calls and fewer than two assistant turns); and
(iv) with non-linear structure, namely multi-agent and branching-tree rollouts.

\paragraph{Stratified sampling.}
We then group the remaining trajectories by (model, task, harness) and discard groups with fewer than 10 trajectories.
To enable cross-task comparison of each model's behavior, we retain only models with groups in at least two task categories, yielding a core set of models with sufficient task overlap.
Finally, we draw a stratified sample for behavioral analysis: from each remaining group, we sample at most 30 trajectories, balancing successful and failed runs when outcomes are available.
This cap bounds the annotation cost per group but does not equalize group sizes; our analyses (\S\ref{sec:stat_tools}) instead account for the imbalance by controlling for task category and harness when testing each feature.
The resulting dataset, used throughout the paper, contains {\numtrajectory} trajectories from {\nummodel} models and {\numharness} harnesses.

\subsection{Feature Construction}
\label{sec:feat_construct}

Psychology has inferred human traits through two channels: behavior, where a trait summarizes the acts a person repeatedly performs~\citep{buss1983act, funder2000riverside}, and language, the lexical hypothesis underlying the Big Five posits that the most salient individual differences are encoded in natural language~\citep{allport1936trait, goldberg1990alternative}, where subsequent work showed that an individual's own word use also reflects their traits~\citep{pennebaker1999linguistic}.
We characterize an LLM through the same two channels: its language yields linguistic features, and its step-by-step actions yield functional features.
For each trajectory, we collect all text generated by the agent, including its reasoning.
We exclude demonstration and scripted events, as well as any text not produced by the agent, such as environment observations, user messages, and tool outputs.

\subsubsection{Functional Features}

We define a step as a single model output, comprising its reasoning, message, and tool calls, and annotate one step of a trajectory at a time.
Annotations follow {\act}~\citep{gao2026interpret}, a recent taxonomy that organizes agent behaviors into 10 groups and 46 sub-actions.
We use GPT-OSS-120B~\citep{gptoss} as the annotator, providing the eight preceding steps and the current observation as read-only context.
For each step, the annotator extracts up to three short quotes (3--15 words each), each capturing a distinct behavioral move, and assigns each quote a group and then a sub-action within that group.
We constrain its output so that every label is a valid taxonomy category and every quote appears verbatim in the step.
Applied to the full corpus, this procedure labels all 947,383 steps across the {\numtrajectory} trajectories at the sub-action level, yielding 1,862,737 quotes (1.97 per step on average).
Agreement with human annotators is reported in Appendix~\ref{app:annotate_compare_human}.

Based on these annotations, we represent each trajectory as a sequence of label sets, defined over the 10 {\act} groups.
We characterize this sequence through four families of properties:
(i) \textit{frequency}~\citep{buss1983act}, the proportion of each behavior, together with overall behavioral diversity and per-step density;
(ii) \textit{co-occurrence}~\citep{bakeman1997observing}, the tendency of two behaviors to appear within the same step;
(iii) \textit{transition}~\citep{chatfield1970analysing}, the probability that a behavior in one step is followed by another in the next, together with the persistence of each behavior across consecutive steps; and
(iv) \textit{stage}~\citep{lashley1951problem}, the distribution of each behavior across the early, middle, and late portions of the trajectory.
In total, each trajectory is described by {\numallfeatureF} functional features; Table~\ref{tab:functional_features} lists the full feature bank.

\subsubsection{Linguistic Features}

Raw text contains formulaic content that carries little information about the agent's linguistic style, such as code blocks, inline code, file paths, URLs, long identifiers, terminal output, and markup.
Because word-frequency measures would count such content as words, we remove it with a small, ordered set of regular expressions.
Table~\ref{tab:prose_cleaning} in the appendix lists the full rule set.
After cleaning, we retain only trajectories with at least 50 words of prose.

We extract linguistic features with LIWC-22 \citep{boyd2022development}, a validated psycholinguistic lexicon that maps English words to categories grouped into higher-level domains.
These categories cover function words (e.g., pronouns, prepositions, and conjunctions), affect (e.g., positive and negative tone), social processes and drives (e.g., social referents and affiliation), cognition (e.g., insight and certitude), and time orientation.
LIWC-22 also provides four summary variables: analytic thinking, clout, authenticity, and emotional tone.
Applying LIWC-22 to each trajectory's cleaned prose yields, for each word category, the percentage of words that belong to it, and for each summary variable, a standardized score between 0 and 100.
We extract {\numallfeatureL} linguistic features for each trajectory.

\subsection{Feature Filtering}
\label{sec:stat_tools}

To ensure that features characterize the model rather than its harness or task, we first apply context adjustment to feature values to remove harness- and task-specific effects, and then apply a three-layer test to identify trait-level features, i.e., features that the same model reproduces stably and that vary sufficiently across models.
The three layers (stability ($\rho_{\text{stab}}$), consistency ($\rho_{\text{con}}$), and discriminability (ICC)) draw on generalizability theory~\citep{shavelson1989generalizability}, and each rules out a distinct way in which a behavior can appear trait-like by accident: a behavior driven by only a few trajectories fails stability, one that separates models on only some tasks fails consistency, and one on which models barely differ fails discriminability.
We justify the threshold for each layer in Appendix~\ref{app:thresholds}.

\paragraph{Context adjustment.}
Our trajectories span multiple models, tasks, and harnesses, so each feature reflects both intrinsic model behavior and artifacts of the evaluation environment.
To make features comparable across models, we remove task and harness effects using a linear mixed-effects model.
We first discard trajectories from harnesses that evaluate only a single model.
For each feature, we then fit the following model via restricted maximum likelihood (REML):
\begin{equation}
y_i = \mu + \alpha_{t(i)} + \beta_{h(i)} + \gamma_{m(i)} + e_i,
\qquad \gamma_m \sim \mathcal{N}(0, \sigma^2_{\text{model}}),
\quad e_i \sim \mathcal{N}(0, \sigma^2),
\end{equation}
where $y_i$ is the feature value of trajectory $i$; $t(i)$, $h(i)$, and $m(i)$ denote its task, harness, and model, respectively; $\mu$ is the intercept; $\alpha_t$ and $\beta_h$ are fixed effects for tasks and harnesses, with reference levels set to zero; $\gamma_m$ is a per-model random intercept; and $e_i$ is the residual.
Including $\gamma_m$ is essential: without it, a harness used predominantly by one model would absorb that model's characteristics into its estimated effect $\hat{\beta}_h$.
We then remove the estimated task and harness offsets while preserving the grand mean of each feature:
\begin{equation}
y^{\text{adj}}_i = y_i - \big(\hat\alpha_{t(i)} + \hat\beta_{h(i)}\big) + \bar c,
\qquad \bar c = \frac{1}{n}\sum_{j=1}^{n}\big(\hat\alpha_{t(j)} + \hat\beta_{h(j)}\big),
\end{equation}
where $n$ is the number of trajectories after filtering.
The adjusted values $y^{\text{adj}}_i$ retain the model effect and the residual, and all downstream analyses are performed on them.

\paragraph{Instance-level stability ($\rho_{\text{stab}}$).}
The first layer uses split-half reliability~\citep{spearman1910correlation, brown1910some} to test whether a feature's model--task cell mean reflects the task as a whole rather than a few trajectories.
For each cell with at least five trajectories, we randomly partition the trajectories into two halves and compute the feature mean within each half.
We then compute the Pearson correlation $r$ between the two sets of half means across all cells and apply the Spearman--Brown correction, $\rho_{\text{stab}} = 2r / (1+r)$, to account for the halved sample size.
We average $\rho_{\text{stab}}$ over ten random splits.
A low $\rho_{\text{stab}}$ indicates that cell means are driven by a few influential trajectories.
We retain features with $\rho_{\text{stab}} \ge 0.7$.

\paragraph{Cross-task consistency ($\rho_{\text{con}}$).}
Whereas the first layer ensures that each cell mean reflects the whole task, the second layer tests whether between-model differences in a feature generalize across tasks rather than reflect task-specific tendencies.
We adapt the cross-situational consistency test from personality psychology~\citep{epstein1979stability, mischel2013personality}.
In the model $\times$ task matrix of cell means, we compute the Pearson correlation between each pair of task columns over the models observed in both tasks, requiring at least four shared models per pair, and average the correlations over all valid pairs.
A $\rho_{\text{con}}$ near zero indicates that between-model differences emerge only in particular tasks.
We retain features with $\rho_{\text{con}} \ge 0.3$.

\paragraph{Model discriminability (ICC).}
The first two layers retain features that are stable within tasks and consistent across tasks; the final layer tests whether these features discriminate between models.
We measure the proportion of variance in the feature attributable to the model using the one-way random-effects intra-class correlation~\citep{shrout1979intraclass}, with models as the grouping factor: $\text{ICC}(1) = \sigma^2_{\text{model}} / (\sigma^2_{\text{model}} + \sigma^2_{\text{within}})$, where $\sigma^2_{\text{model}}$ and $\sigma^2_{\text{within}}$ are the between-model and within-model (i.e., between-trajectory) variance components, estimated by analysis of variance for unequal group sizes.
A high ICC(1) thus indicates that most remaining variation lies between models rather than among trajectories of the same model.
We retain features with $\text{ICC}(1) \ge 0.10$.

\subsection{Factor Analysis}
\label{sec:factor}

To test whether the features that pass the three-layer test are governed by a smaller number of latent dimensions, we perform a factor analysis across models.
Let $\mathbf{Z} \in \mathbb{R}^{n \times p}$ be the data matrix for $n$ models and $p$ features, where $z_{ij}$ is the mean of feature $j$ over the trajectories of model $i$, standardized across models to zero mean and unit variance.
The feature correlation matrix is then $\mathbf{R} = \frac{1}{n-1}\mathbf{Z}^{\!\top}\mathbf{Z}$.
Before extracting factors, we assess the factorability of $\mathbf{R}$ with two standard diagnostics~\citep{dziuban1974correlation}: Bartlett's test of sphericity~\citep{bartlett1937properties}, which tests whether $\mathbf{R}$ differs significantly from the identity matrix, and the Kaiser--Meyer--Olkin (KMO) measure of sampling adequacy~\citep{kaiser1970second, kaiser1974index}.

We adopt the common factor model $\mathbf{R} = \boldsymbol{\Lambda}\boldsymbol{\Lambda}^{\top} + \boldsymbol{\Psi}$, where $\boldsymbol{\Lambda} \in \mathbb{R}^{p \times k}$ is the loading matrix and $\boldsymbol{\Psi} \in \mathbb{R}^{p \times p}$ is the diagonal matrix of uniquenesses.
We initialize the uniquenesses with squared multiple correlations, $\boldsymbol{\Psi}_0 = [\operatorname{diag}(\mathbf{R}^{-1})]^{-1}$, where $\operatorname{diag}(\cdot)$ sets all off-diagonal entries to zero, and form the initial reduced correlation matrix $\mathbf{R}^{*}_0 = \mathbf{R} - \boldsymbol{\Psi}_0$.
We select the number of factors ($k$) by Horn's parallel analysis~\citep{horn1965rationale}: factors are retained sequentially while the $j$-th eigenvalue of $\mathbf{R}^{*}_0$ exceeds the 95th percentile of the $j$-th eigenvalue of reduced correlation matrices computed from 500 simulated standard normal datasets of size $n \times p$~\citep{glorfeld1995improvement}.
We prefer this to the Kaiser criterion (retaining eigenvalues of $\mathbf{R}$ greater than one)~\cite{kaiser1960application}, which tends to over-extract~\citep{zwick1986comparison}.

We estimate $\boldsymbol{\Lambda}$ by iterated principal axis factoring (PAF).
At iteration $t = 0, 1, \dots$, let $\mathbf{V}_t \in \mathbb{R}^{p \times k}$ and $\boldsymbol{\Delta}_t \in \mathbb{R}^{k \times k}$ hold the $k$ leading eigenvectors and eigenvalues of $\mathbf{R}^{*}_t = \mathbf{R} - \boldsymbol{\Psi}_t$, and set $\boldsymbol{\Lambda}_t = \mathbf{V}_t \boldsymbol{\Delta}_t^{1/2}$.
The uniquenesses are then updated as $\boldsymbol{\Psi}_{t+1} = \mathbf{I}_p - \operatorname{diag}(\boldsymbol{\Lambda}_t \boldsymbol{\Lambda}_t^{\top})$.
Starting from the initialization $\boldsymbol{\Psi}_0$, we iterate until $\max_j \lvert \psi_j^{(t+1)} - \psi_j^{(t)} \rvert < \epsilon$, with $\epsilon = 10^{-6}$.

We then rotate the loadings toward simple structure using varimax with Kaiser normalization \citep{kaiser1958varimax}.
Let $\mathbf{H} = [\operatorname{diag}(\boldsymbol{\Lambda}\boldsymbol{\Lambda}^{\top})]^{1/2}$ hold the square roots of the communalities, and let $\mathbf{B} = \mathbf{H}^{-1}\boldsymbol{\Lambda}\mathbf{T}$ denote the row-normalized loadings under an orthogonal rotation $\mathbf{T},\; \mathbf{T}^{\top}\mathbf{T} = \mathbf{I}_k$.
Varimax maximizes the variance of the squared loadings within each factor, summed over factors:
\begin{equation}
\sum_{m=1}^{k} \operatorname{Var}_j\bigl(b_{jm}^{2}\bigr)
= \sum_{m=1}^{k} \Bigl( \mathbb{E}_j\bigl[(b_{jm}^2)^2\bigr] - \bigl(\mathbb{E}_j\bigl[b_{jm}^{2}\bigr]\bigr)^{2} \Bigr)
=\sum_{m=1}^{k} \left[ \frac{1}{p} \sum_{j=1}^{p} b_{jm}^{4}
- \left( \frac{1}{p} \sum_{j=1}^{p} b_{jm}^{2} \right)^{2} \right].
\end{equation}
The rotated loadings are $\boldsymbol{\Lambda}_{\mathrm{rot}} = \mathbf{H}\mathbf{B} = \boldsymbol{\Lambda}\mathbf{T}$.
Since $\mathbf{T}$ is orthogonal, $\boldsymbol{\Lambda}_{\mathrm{rot}}\boldsymbol{\Lambda}_{\mathrm{rot}}^{\top} = \boldsymbol{\Lambda}\boldsymbol{\Lambda}^{\top}$, so the communalities and uniquenesses are unchanged.

Finally, we compute factor scores for each model using Thurstone's regression method~\citep{thurstone1935vectors} with ridge regularization.
The score weights are $\mathbf{W} = (\mathbf{R} + \alpha \mathbf{I}_p)^{-1} \boldsymbol{\Lambda}_{\mathrm{rot}}$, with $\alpha = 0.1$.
The ridge term keeps $\mathbf{W}$ bounded when features are nearly collinear.
We then standardize each column of the raw scores $\mathbf{F} = \mathbf{Z}\mathbf{W}$ to zero mean and unit variance, so that each score reflects a model's standing on that factor relative to the other models in our sample.
\section{Results}
\label{sec:results}

The {\methodname} framework specifies how to derive a low-dimensional, model-attributable behavioral structure from in-the-wild agent trajectories.
We now focus on three research questions (RQs):
(i) \textit{Existence} (\S\ref{sec:stable-trait}): After controlling for task- and harness-induced variance, does agent behavior exhibit a stable, model-attributable structure?
(ii) \textit{Composition} (\S\ref{sec:derived-personality}): What are the stable factors, and which behavioral features load on each?
(iii) \textit{Correlation} (\S\ref{sec:external-correlation}): How do factor scores correlate with external variables, including model release date, task performance, and self-reported trait scores?

\begin{wrapfigure}[14]{r}{0.5\linewidth}
\vspace{-40pt}
\centering
    \includegraphics[width=1.0\linewidth]{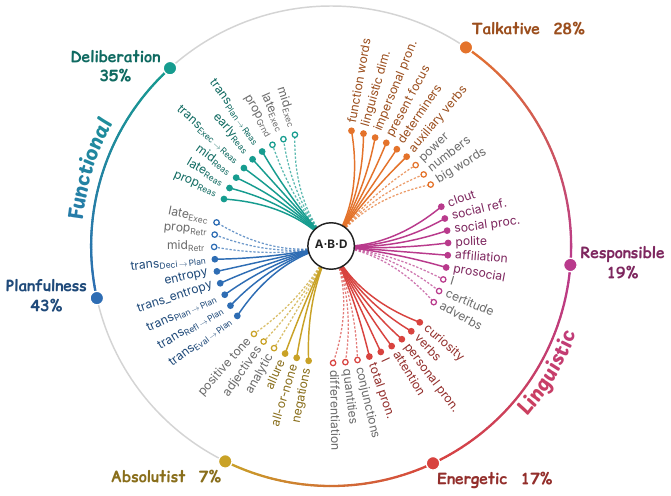}
    \caption{The six behavioral factors and their main features. Each fan is one factor. Leaves are the features with the largest loadings: solid filled dots for positive, dashed hollow dots for negative.}
    \label{fig:factor_radial}
\end{wrapfigure}

\begin{figure}[t]
    \centering
    \includegraphics[width=1.0\linewidth]{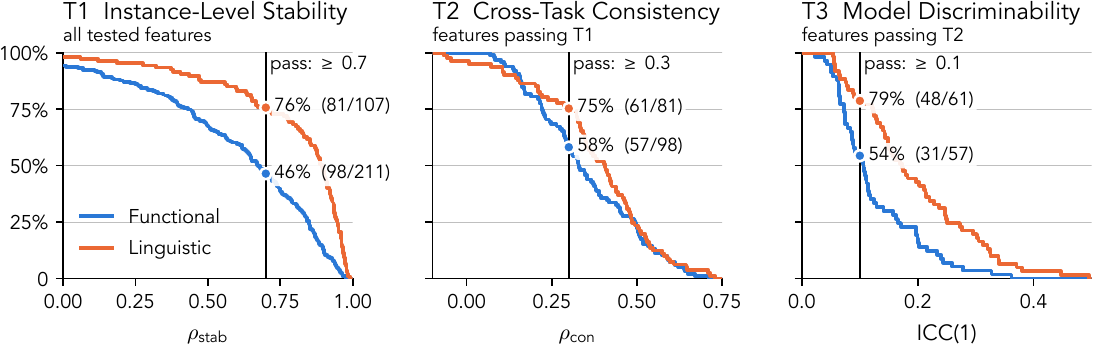}
    \caption{Complementary cumulative distribution of each layer's statistic over 211 functional and 107 linguistic features. Vertical lines mark the thresholds; the height of a curve is the fraction of pass in that layer.}
    \label{fig:trait_layers}
\end{figure}

\subsection{RQ1: Existence of Stable Features}
\label{sec:stable-trait}

As shown in Figure~\ref{fig:trait_layers}, of the {\numallfeature} candidate features we test, 179 are reproducible for the same model on the same task (i.e., they pass the instance-level stability check). Of these, 118 also hold across task types (cross-task consistency check), and {\numfeature} further discriminate between models. These {\numfeature} features comprise {\numfeatureF} functional features (Table~\ref{tab:functional_traits}) and {\numfeatureL} linguistic features (Table~\ref{tab:linguistic_traits}).

\begin{figure}[t]
    \centering
    \begingroup
% Reference-inspired red/orange/blue accents; darker text shades for legibility.
\definecolor{cCurAccent}{HTML}{FF5B65}
\definecolor{cPlanAccent}{HTML}{FFBB5D}
\definecolor{cSocAccent}{HTML}{65AFF6}
\definecolor{cCur}{HTML}{E4515D}
\definecolor{cPlan}{HTML}{C58022}
\definecolor{cSoc}{HTML}{478FCF}
\colorlet{cCurHL}{cCurAccent!18}
\colorlet{cPlanHL}{cPlanAccent!18}
\colorlet{cSocHL}{cSocAccent!18}
% Center each mascot beside its title, aligned at their vertical midpoints.
\newcommand{\profileheading}[2]{%
    \makebox[\linewidth][c]{%
        \raisebox{-0.5\height}{\includegraphics[width=0.40\linewidth]{#1}}%
        \hspace{0.025\linewidth}%
        \small\bfseries\raisebox{-0.5\height}{\shortstack[l]{#2}}}%
}
% Model names are editable here; radar PDFs contain only the plot labels.
\newcommand{\profilemodel}[1]{%
    \makebox[\linewidth][c]{\scriptsize\sffamily #1}%
}
\begin{tabular}{@{}>{\raggedright\arraybackslash}p{0.315\textwidth}@{\hspace{0.0275\textwidth}}>{\raggedright\arraybackslash}p{0.315\textwidth}@{\hspace{0.0275\textwidth}}>{\raggedright\arraybackslash}p{0.315\textwidth}@{}}
\profileheading{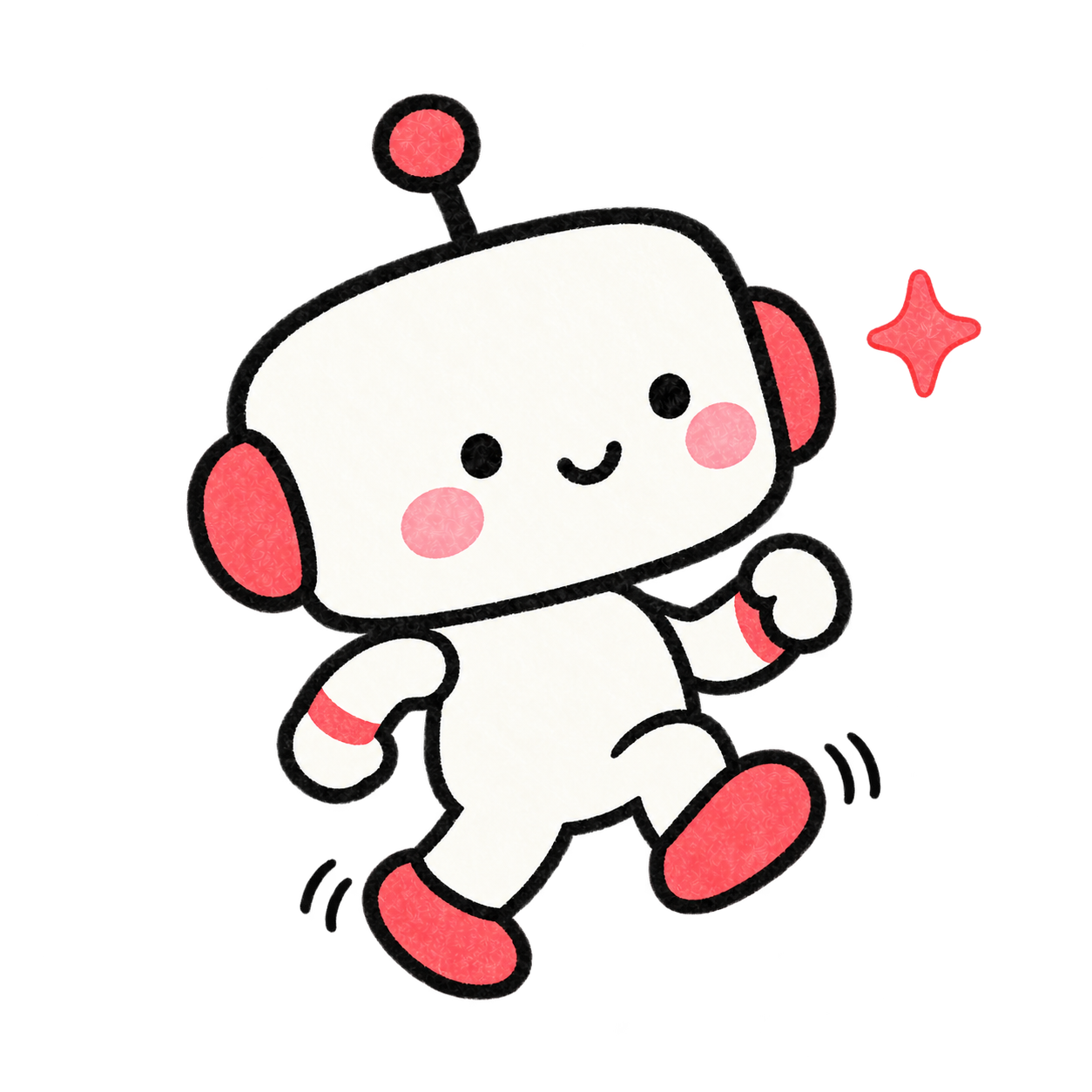}{\textcolor{cCur}{Energetic}} & \profileheading{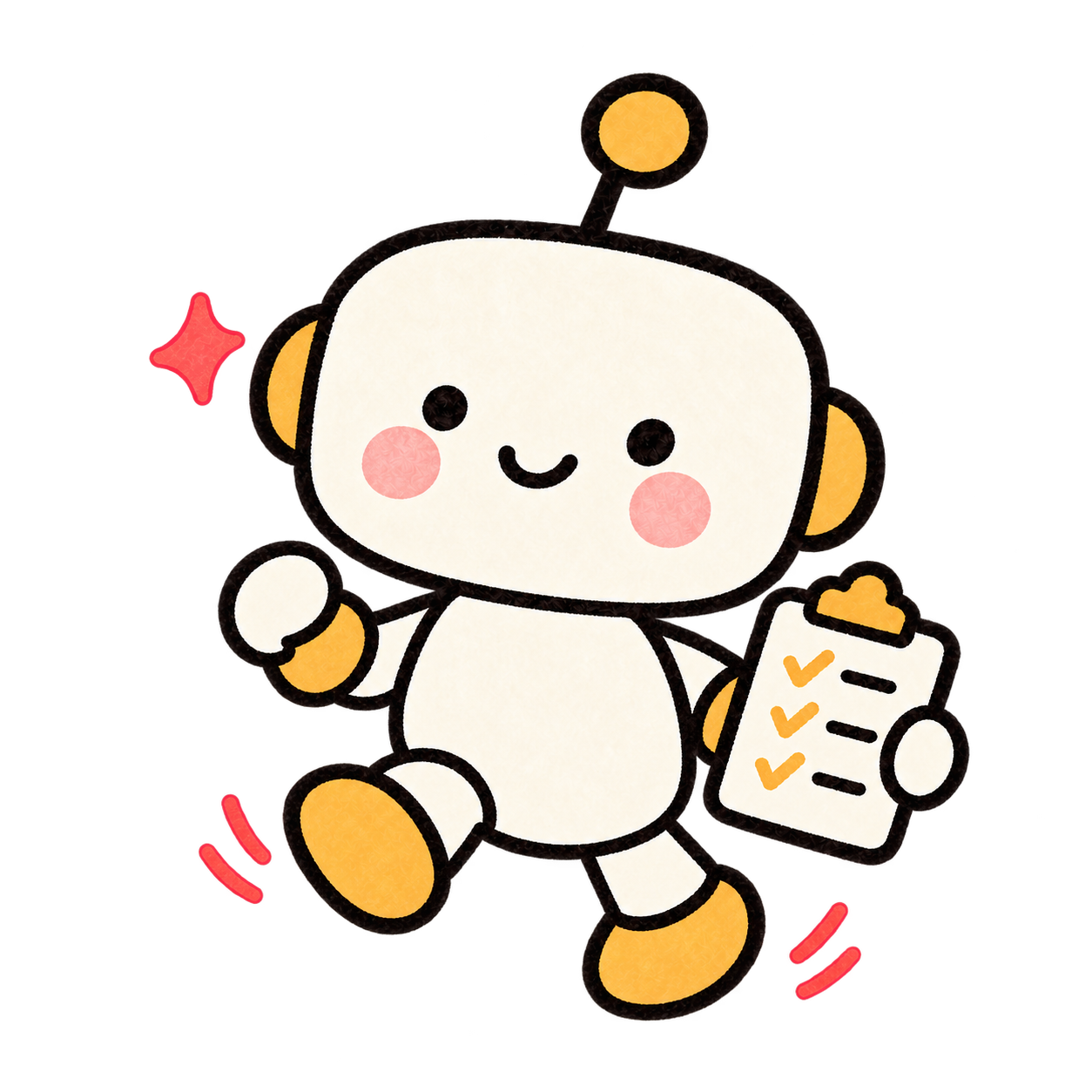}{\textcolor{cPlan}{Planful} +\\\textcolor{cCur}{Energetic}} & \profileheading{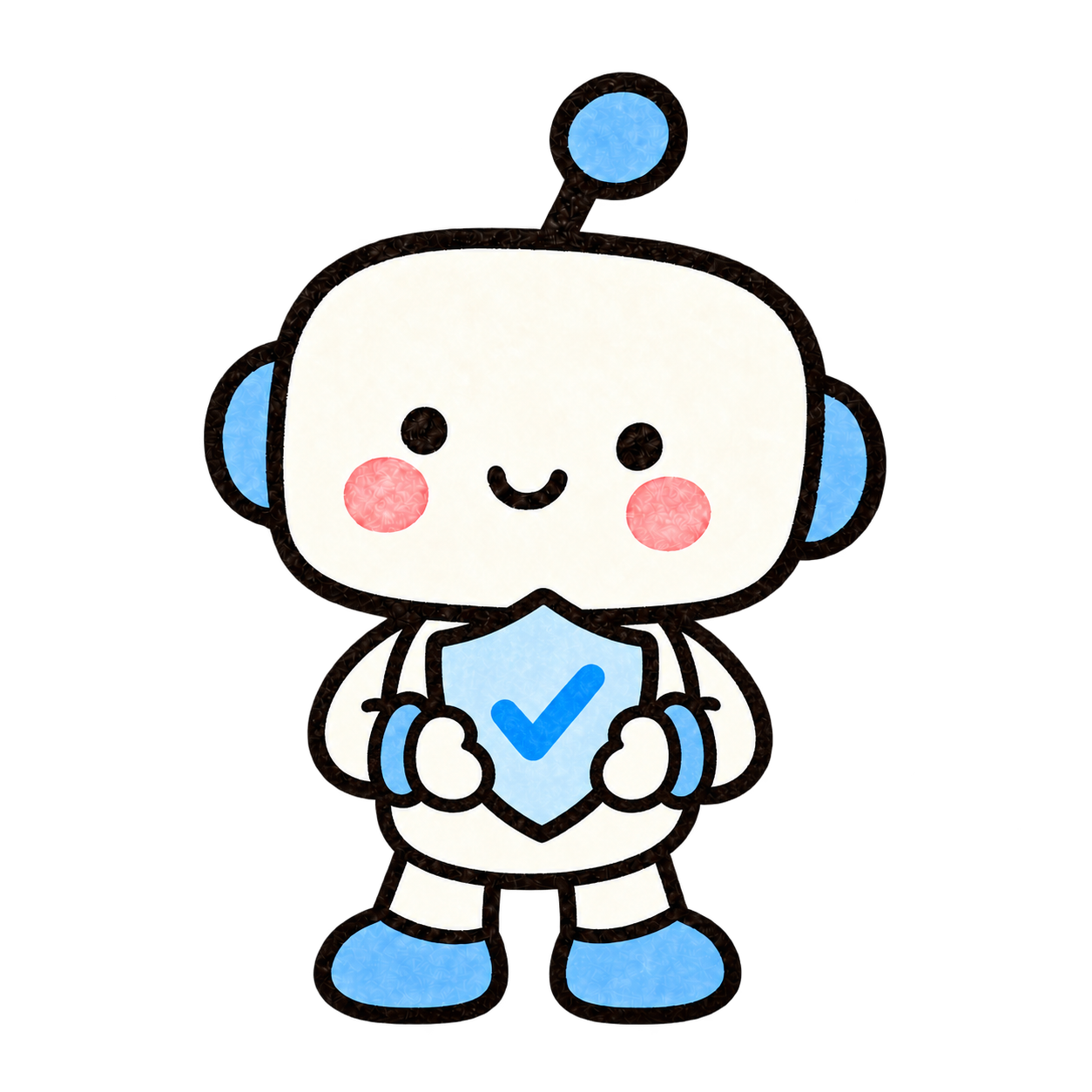}{\textcolor{cSoc}{Responsible}} \\[3pt]
\profilemodel{Claude Sonnet 4.5} & \profilemodel{Qwen3-Coder-480B} & \profilemodel{GPT-5.1} \\[1pt]
\includegraphics[width=\linewidth,trim=0 0 0 22bp,clip]{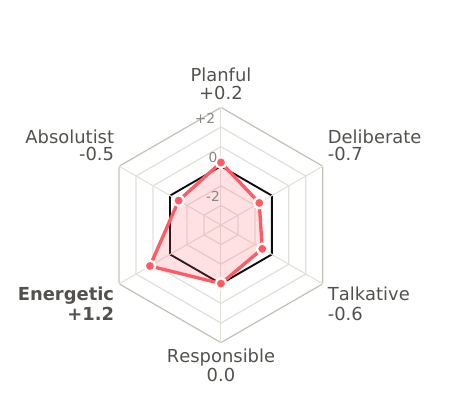} & \includegraphics[width=\linewidth,trim=0 0 0 22bp,clip]{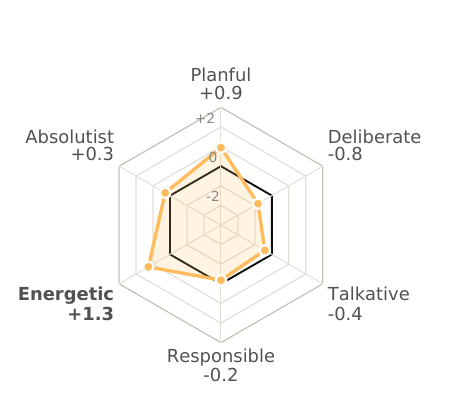} & \includegraphics[width=\linewidth,trim=0 0 0 22bp,clip]{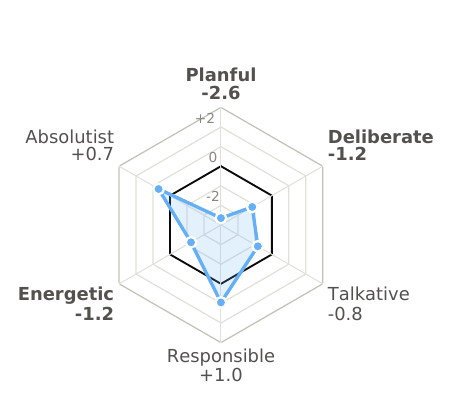} \\[2pt]
{\scriptsize \textcolor{cCur}{\textbf{Energetic $+$1.2}}\par \textbf{[6]} The configure script requires a C compiler, but none is found. I need to install gcc [\ldots] before proceeding with the build.\newline \texttt{\$ apt-\allowbreak{}get install -\allowbreak{}y build-\allowbreak{}essential} \textit{(+1 more command)}\par \textbf{[7]} Build tools including gcc and gcov are now installed. I can proceed with configuring and building SQLite\newline \texttt{\$ CFLAGS=\allowbreak{}"-\allowbreak{}-\allowbreak{}coverage" ./\allowbreak{}configure -\allowbreak{}-\allowbreak{}prefix=\allowbreak{}/\allowbreak{}app/\allowbreak{}sqlite}\par \textbf{[11]} \hlCur{Excellent!} [\ldots] \hlCur{I can see .gcno files were generated} [\ldots] The sqlite3 binary is now in PATH and working \textit{(4 commands)}\par} & {\scriptsize \textcolor{cPlan}{\textbf{Planful $+$0.9}}, \textcolor{cCur}{\textbf{Energetic $+$1.3}}\par \textbf{[12]} The configure script failed because no C compiler was found. \hlPlan{I need to install a C compiler (gcc)} [\ldots] \hlPlan{\emph{Plan:} I'll install gcc and build-essential packages}\newline \texttt{\$ apt-\allowbreak{}get update}\par \textbf{[13]} The package lists have been updated. \hlPlan{Now I need to install gcc and build-essential packages.} \hlPlan{\emph{Plan:} I'll install the build-essential package}\newline \texttt{\$ apt-\allowbreak{}get install -\allowbreak{}y build-\allowbreak{}essential}\par \textbf{[14]} \hlCur{I've successfully installed the build-essential package} [\ldots] \hlPlan{\emph{Plan:} I'll run the configure script again}\newline \texttt{\$ ./\allowbreak{}configure -\allowbreak{}-\allowbreak{}enable-\allowbreak{}gcov -\allowbreak{}-\allowbreak{}prefix=\allowbreak{}/\allowbreak{}usr/\allowbreak{}local}\par} & {\scriptsize \textcolor{cSoc}{\textbf{Responsible $+$1.0}}\par \textbf{[2]} \hlSoc{We are at /app with no work done yet.} Need to unpack the pre-vendored SQLite snapshot \textit{(9 commands)}\par \textbf{[4]} Configure failed because there is no C compiler or make installed [\ldots] \hlSoc{We must first install a build toolchain, then re-run configure and make with coverage flags} [\ldots] \emph{Plan:} 1) \hlSoc{Inspect OS to confirm we can use apt.} [\ldots] 6) Ensure sqlite3 is on PATH\newline \texttt{\$ apt-\allowbreak{}get install -\allowbreak{}y build-\allowbreak{}essential}\newline \texttt{\$ CC=\allowbreak{}gcc CFLAGS=\allowbreak{}'-\allowbreak{}-\allowbreak{}coverage -\allowbreak{}O0' LDFLAGS=\allowbreak{}'-\allowbreak{}-\allowbreak{}coverage' ./\allowbreak{}configure}\newline \texttt{\$ make -\allowbreak{}j"\$(nproc)"} \textit{(+9 more commands)}\par} \\
\end{tabular}
\endgroup

    \caption{Three models on the same task and harness (terminal-bench-2 \texttt{sqlite-with-gcov}, terminus-2). Top: factor scores. Bottom: verbatim excerpts from each trajectory at the same moment, when \texttt{configure} fails for want of a compiler; [$n$] is the step index, \texttt{\$} a command issued in that step, [\ldots] an elision.}
    \label{tab:profiles}
\end{figure}

\subsection{RQ2: Composition of Factors}
\label{sec:derived-personality}

We conduct separate factor analyses for the functional and linguistic features.
The functional feature matrix is suitable for factor analysis (KMO $= 0.89$; Bartlett's $\chi^2(465) = 5713$, $p < 10^{-100}$).
Parallel analysis retains two factors, which jointly explain 78\% of the variance (43\% and 35\%); Table~\ref{tab:loadings_functional} reports the rotated loadings.
The linguistic feature matrix is likewise suitable (KMO $= 0.70$; Bartlett's $\chi^2(1128) = 8923$, $p < 10^{-100}$).
Parallel analysis retains four factors, which jointly explain 71\% of the variance (28\%, 19\%, 17\%, and 7\%); Table~\ref{tab:loadings_linguistic} reports the rotated loadings.
Table~\ref{tab:factor_members} lists the features that load on each of the six factors.
To assess the stability of these factors, we rerun the full pipeline under leave-one-task-out cross-validation, holding out each of the {\numtask} tasks in turn (Appendix~\ref{app:holdout_task_stability}).
As shown in Figure~\ref{fig:factor_radial}, our results demonstrate six observable factors from the agent trajectories.
Figure~\ref{tab:profiles} shows the traits of three models.

\paragraph{Planful.}
A planful model states what it will do next at almost every step and returns to planning after each evaluation, reflection, or decision; its steps also mix more kinds of behavior.
A model low on this factor reads and executes without stating a plan, or plans once at the start and then runs long batches of commands (GPT-5.1 in Figure~\ref{tab:profiles}).
Claude-Sonnet-4.6, Kimi-K3, and Claude-Opus-4.8 are the most planful; GPT-5.1, o3, and o4-Mini are the least, and almost every OpenAI model scores below zero.

\paragraph{Deliberate.}
A deliberate model spends much of each step analyzing what it has just observed before acting, and returns to analysis after each action or plan.
A model low on this factor moves from one tool call to the next with little analysis.
Kimi-K2.7-Code, Llama-3.1-70B, and MiniMax-M2.7 are the most deliberate; Gemini-2.5-Flash, GPT-5-Nano, and GPT-5-Mini are the least.

\paragraph{Talkative.}
A talkative model writes in full sentences, built from function words, articles, pronouns, and auxiliary verbs; the factor measures sentence form, not length.
A model low on this factor writes terse notes dense in numbers and long technical terms.
GPT-4o, the Llama-3 models, and DeepSeek-V2.5 are the most talkative; the GPT-5 Codex models and MiniMax-M2.1 are the least.

\paragraph{Responsible.}
A responsible model writes as if working alongside the reader: \emph{we}, \emph{you}, \emph{please}, \emph{thanks}.
A model low on this factor reports in the first person singular and with certainty (\emph{I}, \emph{actually}).
Claude-3.5-Sonnet and GPT-4o are by far the most responsible; the Qwen3.6 and Qwen3.7 models, Claude-Sonnet-4.6, and Claude-Opus-4.7 are the least.

\paragraph{Energetic.}
An energetic model narrates its own looking and checking, in the first person and with positive interjections: ``let me check'', ``I need to look at'', ``Great, I can see'' (Claude-Sonnet-4.5 in Figure~\ref{tab:profiles}).
A model low on this factor writes flat, qualified description with many adjectives and conjunctions.
DeepSeek-V3.2-Exp, MiniMax-M2.1, GLM-4.6, and the Claude 4.0 to 4.5 models are the most energetic; GPT-5.6, GPT-5.5, Kimi-K3, and Claude-Opus-4.8 are the least.

\paragraph{Absolutist.}
An absolutist model uses many negations and all-or-none words: \emph{not}, \emph{no}, \emph{never}, \emph{all}, \emph{always}.
This is the smallest factor and the least stable under task holdout (Appendix~\ref{app:holdout_task_stability}).
High scores are concentrated in a few models, Gemini-3-Pro, DeepSeek-V3.2, and GPT-OSS-20B; most others score near the mean.

\subsection{RQ3: Correlation with External Metrics}
\label{sec:external-correlation}

\begin{wrapfigure}[23]{r}{0.5\linewidth}
\vspace{-10pt}
\centering
    \includegraphics[width=1.0\linewidth]{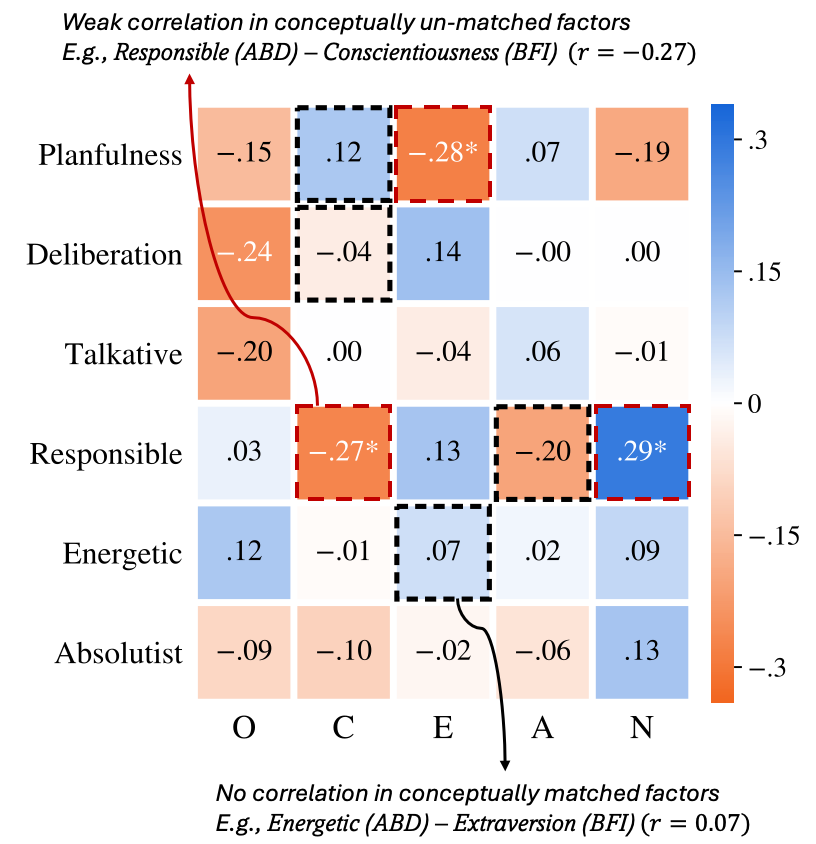}
    \caption{Pearson correlation between factor scores and self-reported Big Five domains.}
    \label{fig:corr_bfi}
\end{wrapfigure}

\paragraph{Correlation with BFI.}
We find that four of our derived factors conceptually match a BFI domain.
\emph{Conscientiousness} in BFI describes a person who makes plans, follows through with them, and thinks before acting.
It matches our \emph{Planful} and \emph{Deliberate} factors, as the former measures whether a model states a plan at each step and the latter measures whether it analyzes before it acts.
\emph{Agreeableness} in BFI describes a person who is cooperative and polite to others.
It matches our \emph{Responsible} factor, which measures whether a model writes as if working with the reader (\emph{we}, \emph{you}, \emph{please}).
\emph{Extraversion} in BFI describes a person who is energetic and full of positive emotion.
It matches our \emph{Energetic} factor, which measures whether a model narrates its own actions with positive interjections.
\emph{Talkative} and \emph{Absolutist} do not match any BFI domain, since the former measures sentence form rather than how much a model says, and the latter has no counterpart in BFI.

Figure~\ref{fig:corr_bfi} presents the correlation between our derived factors and the BFI domains, with the conceptually matched pairs boxed.
Among the matched pairs, the maximum correlation is only $|r| = 0.20$ (Responsible--A, $p = 0.11$), and its sign is even opposite to the match; in the predicted direction the maximum is $r = 0.12$ (Planful--C, $p = 0.34$).
None of the matched pairs is significant, while the three cells with $p < 0.05$ are all unmatched pairs and none of them survives Bonferroni correction.
In comparison, self-reported traits in humans correlate with matched behaviors at around $r = 0.30$~\citep{mischel2013personality}.
Our results suggest that the personality a model reports about itself is largely unrelated to the personality observed from its behavior, i.e., a knowledge--action gap between S-data and B-data.

\paragraph{Correlation with performance and dates.}
We correlate each factor with task performance and model release date (Table~\ref{tab:corr_perf_date}). Regarding performance, only planfulness correlates positively with success ($r = 0.38$, $p < 0.001$), while responsible correlates negatively ($r = -0.31$, $p < 0.01$). In contrast, release date strongly correlates with five factors: newer models are more planful ($r = 0.42$) but less deliberate ($r = -0.24$), talkative ($r = -0.36$), responsible ($r = -0.68$), and energetic ($r = -0.53$). These findings suggest two conclusions. First, planfulness is the only trait that improves over time and directly aids task success. Second, the other generational shifts represent a stylistic drift rather than performance-driven selection. The decline in deliberate, talkative, energetic, and responsible behaviors likely stems from agentic post-training, which rewards terse, plan-driven execution over the conversational style of earlier models. Ultimately, behavioral profiles reveal dimensions of model evolution that task accuracy alone cannot capture.

\begin{table}[t]
\centering\small
\caption{Pearson correlation between factor scores and raw task success (mean normalized outcome over a model's trajectories) and model release date. $^{*}p<0.05$, $^{**}p<0.01$, $^{***}p<0.001$.}
\label{tab:corr_perf_date}
\setlength{\tabcolsep}{5pt}
\newcommand{\pp}{\phantom{-}} % pad positive numbers so digits line up with negatives
\begin{tabular}{lllllll}
\toprule
& \multicolumn{6}{c}{Derived Factors from \methodname} \\
\cmidrule(lr){2-7}
& Planfulness & Deliberation & Talkative & Responsible & Energetic & Absolutist \\
\midrule
Task success  & $\pp0.38^{***}$ & $\pp0.06$   & $-0.00$      & $-0.31^{**}$  & $-0.20$       & $-0.11$ \\
Release date & $\pp0.42^{***}$ & $-0.24^{*}$ & $-0.36^{**}$ & $-0.68^{***}$ & $-0.53^{***}$ & $-0.06$ \\
\bottomrule
\end{tabular}
\end{table}

% \begin{table}[t]
% \centering\small
% \caption{Pearson correlation between factor scores and (i) task performance and (ii) model release date ($n = 78$). Raw success is the mean normalized outcome over a model's trajectories; task-adjusted success is the mean of the within-task-type standardized scores. $^{*}p<0.05$, $^{**}p<0.01$, $^{***}p<0.001$.}
% \label{tab:corr_perf_date}
% \setlength{\tabcolsep}{6pt}
% \begin{tabular}{lccc}
% \toprule
% & \multicolumn{2}{c}{Task performance} & \\
% \cmidrule(lr){2-3}
% Factor & Raw success & Task-adjusted success & Release date \\
% \midrule
% Planfulness  & $0.38^{***}$ & $0.44^{***}$ & $0.42^{***}$ \\
% Deliberation & $0.06$       & $0.05$       & $-0.24^{*}$  \\
% Talkative    & $-0.00$      & $0.06$       & $-0.36^{**}$ \\
% Responsible  & $-0.31^{**}$ & $-0.32^{**}$ & $-0.68^{***}$ \\
% Energetic    & $-0.20$      & $-0.27^{*}$  & $-0.53^{***}$ \\
% Absolutist   & $-0.11$      & $-0.17$      & $-0.06$      \\
% \bottomrule
% \end{tabular}
% \end{table}
\section{Related Work}

\paragraph{AI personality.}
Personality psychology separates what a person says about themselves (S-data) from what informants say about them (I-data)~\citep{funder2015personality}, and LLM personality research has followed both routes. The S-data route administers human inventories to a model~\citep{miotto2022gpt,jiang2023evaluating,huang2024humanity,pellert2024ai,serapio2025psychometric} and reports stable, human-resembling profiles. The I-data route has a judge model infer traits from the target's dialogue, via psychological interviews~\citep{wang2024incharacter}, persona simulation~\citep{wang2025coser}, or multi-observer ratings~\citep{huang2025beyond}, but sees only the conversation elicited for it. Meanwhile, S-data itself has been contested: questionnaire answers show social desirability bias~\citep{salecha2024large}, shift with prompt wording and option order~\citep{gupta2024self}, and fail to reproduce the human factor structure~\citep{suhr2025challenging}.

\paragraph{Knowledge--action gap.}
These critiques share a root: what a model says about itself need not match what it does. A line of work makes this explicit by building a scenario and comparing behavior in it with self-report. TRAIT~\citep{lee2025llms} rewrites inventory items into everyday scenarios; later studies contrast self-reported traits with behavior in social-science experiments~\citep{han2025personality}, values with value-laden decisions~\citep{shen2025mind}, endorsed morality with moral action~\citep{huang2026knowing}, Likert scores with generation probabilities on user queries~\citep{song2025human}, and psychometric facets with matched behavioral paradigms~\citep{yang2026acttraitbench}. The gap persists even with an LLM-native instrument~\citep{contreras2026llm}. Each study designs a new setting and measures the gap there. We need none: we quantify the gap from trajectories agents already produce in the wild, on behaviors that pass a trait test rather than ones chosen in advance.

\paragraph{Bottom-up personality structure.}
A few studies derive a trait structure for models rather than assuming a human one. \citet{mercer2025applying} rerun the HEXACO lexical study on GPT-4 personas, but from human trait adjectives; \citet{jeong2026mti} profile agents on four behavioral axes fixed a priori; \citet{yang2024exploring} steer latent features but score the result on BFI items; \citet{contreras2026llm} factor-analyze responses to purpose-written questionnaire items. None derives the structure from what agents do in deployment. Ours is computed from in-the-wild trajectories and named only afterward.

% A parallel line studies human-like and anthropomorphic behavior directly, in multi-turn interaction and under varied system prompts~\citep{kim2026examining,ibrahim2026multi}, though on fixed conversational settings rather than the open-ended agent tasks where LLMs are increasingly deployed. Evaluation-deployment mismatch: most personality studies are conducted on either a fixed set of questionnaire items or a few predefined tasks.

\paragraph{Trajectory analysis.}
Trajectory analysis examines the process of an agent run rather than only its final outcome. One line of work annotates traces with behavioral or error schemas: \citet{gao2026interpret} label steps to characterize agent behavior, \citet{deshpande2025trail} localize reasoning and execution errors, and \citet{cemri2025multi} categorize failure modes in multi-agent systems. Another line scales trajectory collection; for example, holistic agent leaderboards now release full trajectories alongside task outcomes~\citep{kapoor2026holistic}. These efforts analyze trajectories to interpret, debug, or evaluate individual runs or systems. In contrast, we aggregate many trajectories per model to identify stable, model-distinguishing behavioral traits, recasting trajectory analysis as a measurement of model personality.
\section{Conclusion}

In this paper, we characterize AI personality from behavior rather than from self-report, measuring what a model does and how it writes across a large and heterogeneous set of real agent runs.
We find that models carry stable behavioral traits that hold across tasks and harnesses and give each model a distinct profile.
This profile is a property of the model itself, and it captures a part of an agent that accuracy benchmarks overlook yet users directly experience.
While existing studies rely on carefully designed scenarios to show the knowledge--action gap, we directly quantify this gap using agents' in-the-wild outputs across diverse scenarios.
Our work opens a behavioral axis for evaluating and comparing AI agents that complements task success.
\subsection*{AI use statement}

In this work, we used generative AI tools for the following tasks with required disclosure.
(i)~\emph{Support qualitative data analysis}: functional features in every trajectory were annotated by GPT-OSS-120B under the {\act} taxonomy (\S\ref{sec:feat_construct}); this LLM annotation is a component of our method, and its agreement with a human annotator is reported in Appendix~\ref{app:annotate_compare_human}.
(ii)~\emph{Design or provide feedback on research methodology}: we used Claude Opus and Claude Fable to obtain feedback on the factor analysis pipeline (\S\ref{sec:factor}).
(iii)~\emph{Implement methods}: the analysis pipeline was implemented in Python with the assistance of Claude Opus and Claude Fable.
We have not used generative AI tools to generate synthetic datasets, develop theoretical models or conceptual frameworks, propose or refine hypotheses, assist with translation, clean or reformat datasets, or interpret results.
Formulating mathematical claims, providing critical ingredients for proving mathematical claims, and assisting in the writing of proofs are not applicable to this work, which develops no new mathematical results.
Additionally, we used Claude Opus and Claude Fable for tasks with recommended disclosure: sourcing and searching for publicly available agent trajectories, creating and editing software code, drafting parts of the paper, and editing the paper to improve readability.
We have reviewed all AI-assisted work.
LLM-generated code was verified and tested for correctness by the first author, and all AI-assisted text was checked and revised by the authors.
We take responsibility for the final content of this work, including text, claims, or artifacts produced with the aid of generative AI.

\bibliography{reference, model}
\bibliographystyle{iclr2027_conference}

\clearpage
\appendix
\section{Details of the {\methodname} Framework}

\subsection{Statistics of All Collected Trajectories}

Table~\ref{tab:traj_sources_filtered} lists all {\numbenchmark} sources we collect for trajectories with their assigned task, number of models, harnesses, all (raw) trajectories, and final trajectories selected into the subset.

\begin{table}[h]
\centering
\caption{Raw trajectory sources.
\textbf{Raw} = trajectories collected; \textbf{Subset} = trajectories retained after quality control and stratified sampling.
\textbf{\#Models} and \textbf{\#Harn.} are counts after the pipeline.
For Long-Horizon-Terminal-Bench, only the 503 tasks with a complete trajectory are counted as raw; 698 runs whose log keeps only the last continuation phase were excluded.
The sources vary widely in sizes and tasks, and no single source dominates the subset.}
\label{tab:traj_sources_filtered}
\resizebox{1.0\linewidth}{!}{
\begin{tabular}{p{140pt} p{80pt} r r r r}
    \toprule
    \bf Source & \bf Tasks & \bf \#Models & \bf \#Harn. & \bf \#Raw & \bf \#Subset \\
    \midrule
    GAIA \citep{mialon2024gaia} & research-qa & 3 & 1 & 65{,}915 & 90 \\
    HAL \citep{kapoor2026holistic} & science, swe, dialogue, research-qa, web & 22 & 13 & 58{,}095 & 4{,}876 \\
    APIFlow-Bench \citep{wan2026apiflow} & tool-api & 19 & 1 & 56{,}037 & 570 \\
    Terminal-Bench 2.0 (Community) \citep{merrill2026terminal} & terminal, science, swe, security, ml-eng, other & 31 & 9 & 52{,}272 & 11{,}254 \\
    Multi-SWE-bench \citep{zan2025multi} & swe & 7 & 2 & 27{,}514 & 420 \\
    MCPMark \citep{wu2026mcpmark} & tool-api & 17 & 1 & 11{,}314 & 510 \\
    $\tau^2$-Bench \citep{barres2026tau} & dialogue & 4 & 4 & 10{,}832 & 300 \\
    Terminal-Bench 2.0 (Official) \citep{merrill2026terminal} & terminal, science, swe, security, ml-eng, other & 12 & 12 & 10{,}680 & 3{,}042 \\
    Open General Agent Leaderboard \citep{bandel2026general} & research-qa, swe, dialogue, tool-api & 5 & 5 & 10{,}056 & 2{,}114 \\
    AgentOccam \citep{yang2025agentoccam} & web & 2 & 1 & 7{,}446 & 0 \\
    MALT \citep{parikh2025malt} & research-qa, security, swe, science, ml-eng, web, other & 5 & 1 & 7{,}179 & 893 \\
    Toolathlon \citep{li2026tool} & tool-api & 19 & 1 & 7{,}116 & 570 \\
    LiveClawBench \citep{long2026liveclawbench} & research-qa, terminal, web, swe & 17 & 1 & 6{,}752 & 2{,}009 \\
    OpenHands \citep{wang2025openhands} & swe & 8 & 1 & 5{,}695 & 240 \\
    TheAgentCompany \citep{xu2025theagentcompany} & office & 10 & 3 & 2{,}707 & 330 \\
    $\tau$-Bench \citep{yao2025tau} & dialogue & 1 & 1 & 1{,}980 & 30 \\
    NatureBench \citep{wang2026naturebench} & science & 10 & 2 & 1{,}439 & 282 \\
    OSWorld2.0 \citep{yuan2026osworld2} & gui & 6 & 3 & 1{,}430 & 180 \\
    Cybench \citep{zhang2025cybench} & security & 2 & 1 & 690 & 60 \\
    Long-Horizon-Terminal-Bench \citep{li2026long} & office, ml-eng, swe, other & 2 & 1 & 503 & 33 \\
    WebVoyager \citep{he2024webvoyager} & web & -- & 1 & 15 & 0 \\
    \midrule
    \bf {\numbenchmark} Benchmarks & \bf {\numtask} Tasks & {\nummodel} & {\numharness} & {\numalltrajectory} & {\numtrajectory} \\
    \bottomrule
\end{tabular}
}
\end{table}

\clearpage

\subsection{Filtering Trajectories}
\label{app:funnel}

Table~\ref{tab:funnel} traces the reduction from all collected trajectories to the final subset.
The largest reduction comes from trajectories generated by models outside the target candidate set, most of which originate from large fine-tuning pools.
We then restrict the eligible set to models covering at least two task types, which form the connected core required for the cross-task test; subsample each (model, task, harness) group; and finally remove branching-tree rollouts and runs too short to annotate.

\begin{table}[h]
\centering
\caption{Filtering from all collected trajectories to the final subset. Each discarded trajectory is logged with its exclusion reason for transparency and reproducibility. Eligibility predicates are applied in the listed order, and each trajectory is attributed to the first predicate it fails.}
\label{tab:funnel}
\begin{tabular}{lrr}
    \toprule
    Stage & Dropped & Remaining \\
    \midrule
    Raw collected ({\numbenchmark} sources) &           & {\numalltrajectory} \\
    \midrule
    \quad model not in target candidate set & 63{,}485  & \\
    \quad harness unidentified              & 200       & \\
    \quad task unidentified                 & 1{,}969   & \\
    \quad result record only (no trajectory) & 18{,}294 & \\
    \quad multi-agent (flattened threads)   & 2{,}162   & \\
    \quad fewer than 5 events               & 21{,}181  & \\
    \quad no genuine agent action           & 1{,}308   & \\
    \quad branching-tree rollout            & 1{,}452   & \\
    \quad fewer than 5 annotatable agent steps & 21{,}820 & \\
    \textbf{Eligible}                       &          & 213{,}796 \\
    \midrule
    Groups below floor (138 of 1{,}153 groups) & 520 & \\
    Connectivity: models on $<$2 task types (49 of 129 models) & 33{,}529 & \\
    \textbf{Connected core} (80 models, 960 groups) & & 179{,}747 \\
    \midrule
    Stratified sampling (cap 30 per group)  & 151{,}944 & \\
    \textbf{Final subset}                   &          & \textbf{\numtrajectory} \\
    \bottomrule
\end{tabular}
\end{table}

\clearpage

\subsection{Agreement with Human for Behavioral Annotation}
\label{app:annotate_compare_human}

We used GPT-OSS-120B to annotate functional behaviors of the trajectories.
To validate the annotation results, we compare the LLM-annotated behaviors with human annotations on a randomly sampled subset.
Table~\ref{tab:human-agreement-labels} and \ref{tab:human-agreement-kappa} report the annotation difference.

\begin{table}[h]
\centering
\caption{Step-level agreement between a human annotator and the GPT-OSS-120B annotator on 8 trajectories (56 steps, 131 human and 140 model spans). Each step is reduced to the set of labels its spans carry; Micro-F1 treats the human labels as the reference, Jaccard is the mean per-step overlap of the two label sets.}
\label{tab:human-agreement-labels}
\begin{tabular}{lcccccc}
\toprule
\multirow{2}{*}{Metric} & \multicolumn{2}{c}{Spans per Step} & \multicolumn{2}{c}{Micro-F1} & \multicolumn{2}{c}{Jaccard} \\
\cmidrule(lr){2-3} \cmidrule(lr){4-5} \cmidrule(lr){6-7}
& Human & GPT-OSS-120B & Group & Sub-Action & Group & Sub-Action \\
\midrule
Value & 2.34 & 2.50 & 0.45 & 0.30 & 0.36 & 0.27 \\
\bottomrule
\end{tabular}
\end{table}

\begin{table}[h]
\centering
\caption{Per-group Cohen's $\kappa$ between the human and the GPT-OSS-120B annotator on the same 56 steps. For each group the two annotations are reduced to whether the step carries at least one span of that group; $\kappa$ is the chance-corrected agreement of these binary decisions. Count: steps carrying the group. Learning never occurs on either side ($\kappa$ undefined). Pooled: one $\kappa$ over all 560 step $\times$ group decisions.}
\label{tab:human-agreement-kappa}
\begin{tabular}{lccc}
\toprule
Group & Human & GPT-OSS-120B & $\kappa$ \\
\midrule
Grounding  & 16 & 23 & 0.11 \\
Retrieval  & 11 &  5 & 0.43 \\
Reasoning  &  8 & 18 & 0.14 \\
Planning   & 11 & 22 & 0.22 \\
Evaluate   &  7 &  9 & -0.16 \\
Deciding   &  4 &  3 & 0.24 \\
Executing  & 41 & 34 & 0.25 \\
Reflection &  6 &  2 & -0.06 \\
Learning   &  0 &  0 & -- \\
Memory     & 13 &  0 & 0.00 \\
\midrule
Pooled     &    &    & 0.30 \\
\bottomrule
\end{tabular}
\end{table}

\clearpage

\subsection{Generating Functional Features from Behaviors}

Table~\ref{tab:functional_features} lists all functional features.
Each trajectory's behavior sequence is analyzed along four families: frequency (how often each behavior occurs), co-occurrence (which behaviors share a step), transition (which behavior follows which), and stage (where in the trajectory a behavior falls).
With the indices $g$ and $h$ ranging over the ten {\act} groups, these templates expand to {\numallfeatureF} features.

\begin{table}[h]
\centering
\small
\caption{The functional behavior feature bank. Co-occurrence features are computed within a step, transition and stage features over consecutive steps.}
\label{tab:functional_features}
\begin{tabular}{llp{6.8cm}r}
    \toprule
    Family & Feature & Description & \# \\
    \midrule
    \multirow{3}{*}{Frequency}
    & $\mathrm{prop}_g$    & Share of behavioral spans labeled group $g$. & 10 \\
    & $\mathrm{density}$   & Behavioral spans per agent step. & 1 \\
    & $\mathrm{entropy}$   & Normalized entropy of the group distribution (behavioral variety). & 1 \\
    \midrule
    \multirow{2}{*}{Co-occurrence}
    & $\mathrm{lift}_{g,h}$ & Within-step co-occurrence lift of groups $g$ and $h$: how much more than chance they occur in the same step. & 45 \\
    & $\mathrm{grps\_per\_step}$ & Mean number of distinct groups in a step. & 1 \\
    \midrule
    \multirow{4}{*}{Transition}
    & $\mathrm{trans}_{g\to h}$ & Probability that a step containing group $g$ is followed by a step containing group $h$ (conditional presence; a row need not sum to one). & 100 \\
    & $\mathrm{self\_loop}$ & Probability that a group present in a step is still present in the next step (persistence). & 1 \\
    & $\mathrm{step\_jaccard}$ & Mean Jaccard overlap between the group sets of consecutive steps. & 1 \\
    & $\mathrm{trans\_entropy}$ & Mean normalized entropy of the next-step group distribution. & 1 \\
    \midrule
    \multirow{3}{*}{Stage}
    & $\mathrm{centroid}_g$ & Mean normalized position ($0$ to $1$) of group $g$ across the trajectory. & 10 \\
    & $\mathrm{early}_g,\ \mathrm{mid}_g,\ \mathrm{late}_g$ & Share of group $g$ in the early / middle / late third of the trajectory. & 30 \\
    & $\mathrm{shift}_g$ & Late-third share minus early-third share of group $g$. & 10 \\
    \midrule
    & & \textbf{Total} & \textbf{{\numallfeatureF}} \\
    \bottomrule
\end{tabular}
\end{table}

\clearpage

\subsection{Data Cleaning for Linguistic Features}

Table~\ref{tab:prose_cleaning} lists the process of data cleaning for a better LIWC parsing result.

\begin{table}[h]
\centering
\scriptsize
\caption{Cleaning rules applied to each trajectory's raw text, in the order they run.
Each removal is a single named regular expression: spans matched in the span-stripper block are replaced by a space, and the line-filter block drops whole lines.
Regex flags: \textsuperscript{s} DOTALL, \textsuperscript{m} multiline, \textsuperscript{i} case-insensitive.}
\label{tab:prose_cleaning}
\begin{tabular}{l l p{130pt}}
\toprule
Rule & Pattern & Note \\
\midrule
\multicolumn{3}{l}{\emph{(a) Message level}}\\
JSON prose & \emph{message parses as a JSON dict} & keep only fields \texttt{analysis}, \texttt{plan}, \texttt{thought}, \texttt{reasoning}, \texttt{reflection}; drop the rest (e.g.\ command payloads) \\
\midrule
\multicolumn{3}{l}{\emph{(b) Span strippers (matched span $\rightarrow$ space)}}\\
Fenced code    & \verb!```.*?```!\textsuperscript{s}       & fenced code blocks \\
Dangling fence & \verb!```.*!\textsuperscript{s}           & an unclosed trailing code fence \\
Think tags     & \verb!</?(mm:)?think>!\textsuperscript{i} & reasoning-wrapper tags (e.g.\ MiniMax \texttt{<mm:think>}) \\
Inline markup  & \verb!<[^>\n]{1,200}>!                    & inline HTML / XML tags \\
Inline code    & \verb!`[^`\n]+`!                          & backtick code spans \\
URL            & \verb!https?://\S+|www\.\S+!              & web links \\
File path      & \verb!(?:\.{0,2}/)?(?:[\w.\-]+/){2,}[\w.\-]*! & slash paths of $\ge\!2$ segments \\
Long hex       & \verb!\b[0-9a-fA-F]{16,}\b!               & hashes / ids ($\ge\!16$ hex chars) \\
Long number    & \verb!\b\d{7,}\b!                         & digit runs of $\ge\!7$ \\
\midrule
\multicolumn{3}{l}{\emph{(c) Line filters (whole line dropped)}}\\
Prompt line    & \verb!^\s*(root@|$ |# |>>> |\.\.\. )!\textsuperscript{m} & lines starting with a shell or REPL prompt \\
Non-prose line & \emph{letter ratio $<0.55$ or $<2$ words} & tables, hexdumps, code fragments, lone tokens \\
\midrule
\multicolumn{3}{l}{\emph{(d) Normalization}}\\
Markdown markers & \verb!^\s*[#>\-\*\d\.\)]+\s*!\textsuperscript{m} & leading heading / bullet / list / quote markers \\
Whitespace     & \verb!\s+! $\rightarrow$ space             & collapse runs of whitespace \\
\bottomrule
\end{tabular}
\end{table}

\clearpage

\section{More Results}

\subsection{Three-Layer Test: Score Distribution}
\label{app:thresholds}

Figure~\ref{fig:trait_layers} shows the complementary cumulative distribution of each layer's statistic over all tested features, separately for functional and linguistic features. Because a feature passes a layer when its statistic meets or exceeds the layer's threshold, the height of each curve at the dashed line gives the fraction of features that pass that layer.

We set the thresholds following psychometric convention and the data distribution.
For instance-level stability, $0.70$ is the minimum reliability conventionally recommended by \citet{nunnally1978psychometric}.
For cross-task consistency, correlations in human personality research rarely exceed $0.30$, whether between a trait measure and behavior in a specific situation or between behaviors across situations~\citep{mischel2013personality}.
We therefore require a feature to be at least as consistent across tasks as human traits typically are across situations.
For model discriminability, ICC values are concentrated near zero in our data, and a threshold of $0.10$, a medium effect for a variance-partition coefficient~\citep{lebreton2008answers}, separates features with non-negligible between-model variance from the bulk with almost none.

\subsection{Features Passing the Three-Layer Test}
\label{app:three_layer_test_for_trait}

\begin{table}[h]
\centering
\small
\setlength{\tabcolsep}{4pt}
\caption{The {\numfeatureF} functional features that pass all three tests, grouped by the feature families of Table~\ref{tab:functional_features}.
{\act} groups are abbreviated in subscripts: Plan = Planning, Reas = Reasoning, Exec = Executing, Eval = Evaluate, Refl = Reflection, Grnd = Grounding, Retr = Retrieval, Deci = Deciding.
$\rho_{\text{stab}}$: split-half reliability within a task; $\rho_{\text{con}}$: cross-task consistency; ICC: between-model discriminability.}
\label{tab:functional_traits}
\begin{minipage}[t]{0.49\linewidth}\centering
\begin{tabular}{llrrr}
\toprule
Family & Feature & $\rho_{\text{stab}}$ & $\rho_{\text{con}}$ & ICC \\
\midrule
\multirow{14}{*}{Transition}
& $\mathrm{trans\_entropy}$ & 0.98 & 0.64 & 0.36 \\
& $\mathrm{trans}_{\mathrm{Plan}\to\mathrm{Plan}}$ & 0.95 & 0.65 & 0.24 \\
& $\mathrm{trans}_{\mathrm{Reas}\to\mathrm{Plan}}$ & 0.93 & 0.50 & 0.20 \\
& $\mathrm{trans}_{\mathrm{Reas}\to\mathrm{Reas}}$ & 0.94 & 0.55 & 0.20 \\
& $\mathrm{trans}_{\mathrm{Exec}\to\mathrm{Reas}}$ & 0.94 & 0.55 & 0.20 \\
& $\mathrm{trans}_{\mathrm{Plan}\to\mathrm{Reas}}$ & 0.95 & 0.52 & 0.20 \\
& $\mathrm{trans}_{\mathrm{Exec}\to\mathrm{Plan}}$ & 0.94 & 0.59 & 0.20 \\
& $\mathrm{trans}_{\mathrm{Eval}\to\mathrm{Plan}}$ & 0.89 & 0.53 & 0.14 \\
& $\mathrm{trans}_{\mathrm{Refl}\to\mathrm{Plan}}$ & 0.88 & 0.60 & 0.13 \\
& $\mathrm{trans}_{\mathrm{Refl}\to\mathrm{Exec}}$ & 0.87 & 0.62 & 0.11 \\
& $\mathrm{trans}_{\mathrm{Deci}\to\mathrm{Plan}}$ & 0.85 & 0.46 & 0.11 \\
& $\mathrm{trans}_{\mathrm{Refl}\to\mathrm{Reas}}$ & 0.86 & 0.51 & 0.11 \\
& $\mathrm{trans}_{\mathrm{Eval}\to\mathrm{Reas}}$ & 0.85 & 0.48 & 0.11 \\
& $\mathrm{trans}_{\mathrm{Plan}\to\mathrm{Exec}}$ & 0.90 & 0.38 & 0.11 \\[\dimexpr\aboverulesep+\lightrulewidth+\belowrulesep\relax]
\midrule
Co-occurrence
& $\mathrm{grps\_per\_step}$ & 0.98 & 0.59 & 0.33 \\[\dimexpr\aboverulesep+\lightrulewidth+\belowrulesep\relax]
\bottomrule
\end{tabular}
\end{minipage}\hfill
\begin{minipage}[t]{0.49\linewidth}\centering
\begin{tabular}{llrrr}
\toprule
Family & Feature & $\rho_{\text{stab}}$ & $\rho_{\text{con}}$ & ICC \\
\midrule
\multirow{7}{*}{Frequency}
& $\mathrm{prop}_{\mathrm{Reas}}$ & 0.96 & 0.49 & 0.28 \\
& $\mathrm{entropy}$ & 0.96 & 0.55 & 0.26 \\
& $\mathrm{prop}_{\mathrm{Plan}}$ & 0.96 & 0.52 & 0.24 \\
& $\mathrm{density}$ & 0.96 & 0.51 & 0.22 \\
& $\mathrm{prop}_{\mathrm{Grnd}}$ & 0.95 & 0.33 & 0.16 \\
& $\mathrm{prop}_{\mathrm{Refl}}$ & 0.90 & 0.71 & 0.12 \\
& $\mathrm{prop}_{\mathrm{Retr}}$ & 0.94 & 0.31 & 0.11 \\
\midrule
\multirow{9}{*}{Stage}
& $\mathrm{early}_{\mathrm{Reas}}$ & 0.95 & 0.51 & 0.23 \\
& $\mathrm{early}_{\mathrm{Plan}}$ & 0.94 & 0.39 & 0.17 \\
& $\mathrm{mid}_{\mathrm{Reas}}$ & 0.93 & 0.50 & 0.17 \\
& $\mathrm{mid}_{\mathrm{Plan}}$ & 0.92 & 0.57 & 0.16 \\
& $\mathrm{late}_{\mathrm{Exec}}$ & 0.90 & 0.32 & 0.12 \\
& $\mathrm{late}_{\mathrm{Reas}}$ & 0.90 & 0.30 & 0.11 \\
& $\mathrm{mid}_{\mathrm{Exec}}$ & 0.90 & 0.31 & 0.11 \\
& $\mathrm{late}_{\mathrm{Plan}}$ & 0.88 & 0.49 & 0.11 \\
& $\mathrm{mid}_{\mathrm{Retr}}$ & 0.91 & 0.33 & 0.10 \\
\bottomrule
\end{tabular}
\end{minipage}
\end{table}

\begin{table}[h]
\centering
\small
\setlength{\tabcolsep}{4pt}
\caption{The {\numfeatureL} linguistic features that pass all three tests.
Features are grouped by LIWC-22's own output-header domains (\textit{Function words} abbreviates LIWC's \textit{Linguistic Dimensions} header, \textit{Summary} its summary variables, and \textit{Motives} is our label for the header-less \texttt{need}/\dots/\texttt{allure} block).
Categories are scored word-level on the full cleaned corpus.
$\rho_{\text{stab}}$: split-half reliability within a task; $\rho_{\text{con}}$: cross-task consistency; ICC: between-model discriminability.}
\label{tab:linguistic_traits}
\begin{minipage}[t]{0.49\linewidth}\centering
\begin{tabular}{llrrr}
\toprule
Domain & LIWC Cat. & $\rho_{\text{stab}}$ & $\rho_{\text{con}}$ & ICC \\
\midrule
\multirow{19}{*}{\shortstack[l]{Function\\ words}}
& we & 0.99 & 0.71 & 0.49 \\
& function words & 0.98 & 0.46 & 0.37 \\
& articles & 0.98 & 0.46 & 0.34 \\
& linguistic dim. & 0.97 & 0.42 & 0.34 \\
& personal pron. & 0.97 & 0.59 & 0.32 \\
& total pron. & 0.97 & 0.51 & 0.32 \\
& I & 0.97 & 0.55 & 0.32 \\
& determiners & 0.98 & 0.41 & 0.30 \\
& auxiliary verbs & 0.97 & 0.53 & 0.27 \\
& verbs & 0.96 & 0.40 & 0.25 \\
& adverbs & 0.95 & 0.44 & 0.25 \\
& conjunctions & 0.95 & 0.58 & 0.23 \\
& you & 0.97 & 0.42 & 0.22 \\
& impersonal pron. & 0.95 & 0.37 & 0.20 \\
& prepositions & 0.95 & 0.52 & 0.18 \\
& adjectives & 0.94 & 0.34 & 0.18 \\
& numbers & 0.94 & 0.49 & 0.16 \\
& quantities & 0.91 & 0.36 & 0.15 \\
& negations & 0.96 & 0.53 & 0.14 \\
\midrule
\multirow{5}{*}{Summary}
& clout & 0.98 & 0.65 & 0.38 \\
& big words & 0.97 & 0.37 & 0.25 \\
& analytic & 0.93 & 0.44 & 0.17 \\
& emotional tone & 0.89 & 0.51 & 0.15 \\
& authentic & 0.91 & 0.32 & 0.11 \\
\midrule
Perception
& attention & 0.92 & 0.35 & 0.15 \\[\dimexpr\aboverulesep+\lightrulewidth+\belowrulesep\relax]
\bottomrule
\end{tabular}
\end{minipage}\hfill
\begin{minipage}[t]{0.49\linewidth}\centering
\begin{tabular}{llrrr}
\toprule
Domain & LIWC Cat. & $\rho_{\text{stab}}$ & $\rho_{\text{con}}$ & ICC \\
\midrule
\multirow{2}{*}{Affect}
& positive emotion & 0.89 & 0.47 & 0.13 \\
& positive tone & 0.88 & 0.48 & 0.12 \\
\midrule
\multirow{4}{*}{Social}
& social ref. & 0.96 & 0.65 & 0.31 \\
& social proc. & 0.93 & 0.54 & 0.17 \\
& prosocial & 0.90 & 0.57 & 0.14 \\
& polite & 0.91 & 0.48 & 0.13 \\
\midrule
\multirow{4}{*}{Drives}
& affiliation & 0.98 & 0.73 & 0.45 \\
& power & 0.98 & 0.33 & 0.29 \\
& drives (total) & 0.96 & 0.49 & 0.21 \\
& achievement & 0.92 & 0.47 & 0.17 \\
\midrule
\multirow{5}{*}{Cognition}
& certitude & 0.93 & 0.60 & 0.21 \\
& insight & 0.93 & 0.54 & 0.20 \\
& discrepancy & 0.93 & 0.42 & 0.14 \\
& differentiation & 0.90 & 0.44 & 0.12 \\
& all-or-none & 0.96 & 0.38 & 0.12 \\
\midrule
\multirow{2}{*}{Lifestyle}
& work & 0.96 & 0.33 & 0.25 \\
& lifestyle (total) & 0.95 & 0.42 & 0.21 \\
\midrule
\multirow{2}{*}{\shortstack[l]{Time\\ orient.}}
& present focus & 0.97 & 0.41 & 0.30 \\
& future focus & 0.96 & 0.71 & 0.28 \\
\midrule
\multirow{3}{*}{Motives}
& curiosity & 0.94 & 0.61 & 0.24 \\
& need & 0.92 & 0.48 & 0.16 \\
& allure & 0.94 & 0.47 & 0.15 \\
\midrule
Conversation
& assent & 0.95 & 0.48 & 0.14 \\
\bottomrule
\end{tabular}
\end{minipage}
\end{table}

\clearpage

\subsection{Loadings of Features}

\begin{table}[h]
\centering
\caption{Functional family: varimax-rotated loadings of the {\numfeatureF} functional features on the two factors retained by parallel analysis (KMO $= 0.89$). Loadings with $|\lambda| \ge 0.40$ are marked in bold; $h^2$ is the communality. Traits are ordered by their primary factor and loading.}
\label{tab:loadings_functional}
\begin{tabular}{lccc}
\toprule
Trait & F1 & F2 & $h^2$ \\
\midrule
$\mathrm{trans}_{\mathrm{Eval}\to\mathrm{Plan}}$ & $\mathbf{0.90}$ & 0.35 & 0.94 \\
$\mathrm{trans}_{\mathrm{Refl}\to\mathrm{Plan}}$ & $\mathbf{0.88}$ & 0.27 & 0.85 \\
$\mathrm{trans}_{\mathrm{Plan}\to\mathrm{Plan}}$ & $\mathbf{0.85}$ & $\mathbf{0.46}$ & 0.94 \\
$\mathrm{trans\_entropy}$ & $\mathbf{0.85}$ & 0.30 & 0.81 \\
$\mathrm{entropy}$ & $\mathbf{0.84}$ & 0.35 & 0.83 \\
$\mathrm{trans}_{\mathrm{Deci}\to\mathrm{Plan}}$ & $\mathbf{0.81}$ & 0.29 & 0.74 \\
$\mathrm{trans}_{\mathrm{Refl}\to\mathrm{Exec}}$ & $\mathbf{0.80}$ & 0.16 & 0.67 \\
$\mathrm{trans}_{\mathrm{Reas}\to\mathrm{Plan}}$ & $\mathbf{0.80}$ & $\mathbf{0.54}$ & 0.92 \\
$\mathrm{grps\_per\_step}$ & $\mathbf{0.79}$ & $\mathbf{0.55}$ & 0.92 \\
$\mathrm{prop}_{\mathrm{Plan}}$ & $\mathbf{0.78}$ & $\mathbf{0.53}$ & 0.89 \\
$\mathrm{mid}_{\mathrm{Plan}}$ & $\mathbf{0.74}$ & $\mathbf{0.53}$ & 0.84 \\
$\mathrm{late}_{\mathrm{Plan}}$ & $\mathbf{0.74}$ & $\mathbf{0.50}$ & 0.81 \\
$\mathrm{mid}_{\mathrm{Retr}}$ & $\mathbf{-0.74}$ & $-0.27$ & 0.62 \\
$\mathrm{early}_{\mathrm{Plan}}$ & $\mathbf{0.74}$ & $\mathbf{0.53}$ & 0.82 \\
$\mathrm{trans}_{\mathrm{Refl}\to\mathrm{Reas}}$ & $\mathbf{0.71}$ & $\mathbf{0.54}$ & 0.80 \\
$\mathrm{prop}_{\mathrm{Retr}}$ & $\mathbf{-0.67}$ & $-0.30$ & 0.54 \\
$\mathrm{trans}_{\mathrm{Exec}\to\mathrm{Plan}}$ & $\mathbf{0.66}$ & $\mathbf{0.59}$ & 0.78 \\
$\mathrm{density}$ & $\mathbf{0.62}$ & $\mathbf{0.45}$ & 0.59 \\
$\mathrm{prop}_{\mathrm{Refl}}$ & $\mathbf{0.55}$ & 0.21 & 0.35 \\
$\mathrm{trans}_{\mathrm{Plan}\to\mathrm{Exec}}$ & 0.32 & 0.26 & 0.17 \\
$\mathrm{prop}_{\mathrm{Reas}}$ & 0.30 & $\mathbf{0.95}$ & 0.99 \\
$\mathrm{late}_{\mathrm{Reas}}$ & 0.16 & $\mathbf{0.94}$ & 0.92 \\
$\mathrm{mid}_{\mathrm{Reas}}$ & 0.32 & $\mathbf{0.92}$ & 0.95 \\
$\mathrm{trans}_{\mathrm{Exec}\to\mathrm{Reas}}$ & 0.36 & $\mathbf{0.92}$ & 0.97 \\
$\mathrm{early}_{\mathrm{Reas}}$ & 0.37 & $\mathbf{0.88}$ & 0.91 \\
$\mathrm{trans}_{\mathrm{Plan}\to\mathrm{Reas}}$ & $\mathbf{0.47}$ & $\mathbf{0.84}$ & 0.93 \\
$\mathrm{trans}_{\mathrm{Reas}\to\mathrm{Reas}}$ & $\mathbf{0.51}$ & $\mathbf{0.81}$ & 0.91 \\
$\mathrm{trans}_{\mathrm{Eval}\to\mathrm{Reas}}$ & $\mathbf{0.55}$ & $\mathbf{0.73}$ & 0.84 \\
$\mathrm{prop}_{\mathrm{Grnd}}$ & $-0.39$ & $\mathbf{-0.72}$ & 0.67 \\
$\mathrm{late}_{\mathrm{Exec}}$ & $\mathbf{-0.60}$ & $\mathbf{-0.61}$ & 0.72 \\
$\mathrm{mid}_{\mathrm{Exec}}$ & $\mathbf{-0.47}$ & $\mathbf{-0.58}$ & 0.56 \\
\midrule
Variance explained & 43\% & 35\% & \\
\bottomrule
\end{tabular}
\end{table}

\begin{table}[h]
\centering
\caption{Linguistic family: varimax-rotated loadings of the {\numfeatureL} linguistic features on the four factors retained by parallel analysis (KMO $= 0.70$). Loadings with $|\lambda| \ge 0.40$ are marked in bold; $h^2$ is the communality. Traits are ordered by their primary factor and loading.}
\label{tab:loadings_linguistic}
\begin{tabular}{lccccc}
\toprule
Trait & F1 & F2 & F3 & F4 & $h^2$ \\
\midrule
function words & $\mathbf{0.95}$ & 0.08 & 0.19 & 0.01 & 0.94 \\
linguistic dim. & $\mathbf{0.93}$ & $-0.00$ & 0.30 & $-0.04$ & 0.96 \\
impersonal pron. & $\mathbf{0.91}$ & $-0.10$ & 0.12 & $-0.14$ & 0.88 \\
present focus & $\mathbf{0.89}$ & 0.03 & 0.12 & $-0.06$ & 0.81 \\
determiners & $\mathbf{0.88}$ & 0.12 & 0.15 & $-0.00$ & 0.81 \\
auxiliary verbs & $\mathbf{0.88}$ & 0.14 & 0.13 & $-0.01$ & 0.81 \\
articles & $\mathbf{0.87}$ & 0.09 & 0.16 & $-0.14$ & 0.80 \\
power & $\mathbf{-0.85}$ & 0.24 & 0.26 & 0.17 & 0.87 \\
numbers & $\mathbf{-0.83}$ & $-0.20$ & $-0.12$ & $-0.03$ & 0.74 \\
big words & $\mathbf{-0.73}$ & $\mathbf{0.42}$ & 0.21 & $-0.05$ & 0.76 \\
lifestyle (total) & $\mathbf{-0.72}$ & 0.32 & $\mathbf{0.48}$ & 0.16 & 0.87 \\
total pron. & $\mathbf{0.72}$ & 0.19 & $\mathbf{0.61}$ & $-0.01$ & 0.92 \\
adverbs & $\mathbf{0.71}$ & $\mathbf{-0.48}$ & $-0.21$ & $-0.04$ & 0.78 \\
work & $\mathbf{-0.69}$ & 0.32 & $\mathbf{0.52}$ & 0.18 & 0.89 \\
prepositions & $\mathbf{0.66}$ & 0.18 & 0.30 & $-0.04$ & 0.55 \\
achievement & $\mathbf{0.52}$ & $-0.40$ & 0.25 & $-0.37$ & 0.63 \\
discrepancy & $\mathbf{0.48}$ & $-0.07$ & 0.23 & 0.28 & 0.37 \\
assent & $-0.28$ & $-0.21$ & $-0.22$ & 0.04 & 0.17 \\
clout & 0.14 & $\mathbf{0.92}$ & 0.18 & 0.12 & 0.92 \\
social ref. & 0.27 & $\mathbf{0.87}$ & 0.27 & 0.21 & 0.95 \\
social proc. & $-0.28$ & $\mathbf{0.83}$ & 0.12 & 0.30 & 0.88 \\
polite & $-0.16$ & $\mathbf{0.83}$ & 0.15 & $-0.02$ & 0.73 \\
affiliation & $\mathbf{0.41}$ & $\mathbf{0.83}$ & 0.13 & 0.15 & 0.90 \\
prosocial & $-0.09$ & $\mathbf{0.80}$ & 0.35 & $-0.07$ & 0.78 \\
we & $\mathbf{0.46}$ & $\mathbf{0.79}$ & 0.09 & 0.16 & 0.87 \\
you & $-0.11$ & $\mathbf{0.71}$ & $\mathbf{0.60}$ & 0.11 & 0.88 \\
I & 0.23 & $\mathbf{-0.62}$ & $\mathbf{0.62}$ & $-0.15$ & 0.85 \\
certitude & 0.24 & $\mathbf{-0.61}$ & $-0.33$ & $-0.25$ & 0.60 \\
drives (total) & $\mathbf{-0.52}$ & $\mathbf{0.61}$ & $\mathbf{0.45}$ & 0.12 & 0.85 \\
authentic & 0.34 & $-0.37$ & $-0.01$ & $-0.27$ & 0.32 \\
curiosity & 0.14 & 0.38 & $\mathbf{0.86}$ & 0.10 & 0.91 \\
verbs & $\mathbf{0.41}$ & 0.06 & $\mathbf{0.77}$ & 0.09 & 0.77 \\
personal pron. & $\mathbf{0.48}$ & 0.31 & $\mathbf{0.75}$ & 0.06 & 0.89 \\
attention & 0.14 & $\mathbf{0.42}$ & $\mathbf{0.66}$ & 0.15 & 0.66 \\
positive emotion & 0.15 & 0.11 & $\mathbf{0.60}$ & $-0.15$ & 0.41 \\
emotional tone & 0.34 & 0.32 & $\mathbf{0.59}$ & $-0.26$ & 0.63 \\
conjunctions & 0.28 & 0.17 & $\mathbf{-0.58}$ & $-0.02$ & 0.45 \\
insight & 0.16 & 0.33 & $\mathbf{0.57}$ & $-0.11$ & 0.47 \\
quantities & 0.29 & $\mathbf{-0.45}$ & $\mathbf{-0.55}$ & $-0.39$ & 0.74 \\
differentiation & $\mathbf{0.46}$ & $-0.18$ & $\mathbf{-0.52}$ & 0.10 & 0.52 \\
adjectives & $\mathbf{0.46}$ & $\mathbf{-0.40}$ & $\mathbf{-0.51}$ & $\mathbf{-0.43}$ & 0.82 \\
need & $-0.02$ & 0.28 & $\mathbf{0.48}$ & 0.37 & 0.45 \\
positive tone & 0.29 & 0.22 & $\mathbf{0.45}$ & $\mathbf{-0.40}$ & 0.50 \\
future focus & 0.24 & $\mathbf{0.41}$ & $\mathbf{0.44}$ & $-0.08$ & 0.43 \\
negations & $-0.08$ & 0.26 & $-0.09$ & $\mathbf{0.89}$ & 0.88 \\
all-or-none & $-0.29$ & 0.14 & $-0.02$ & $\mathbf{0.74}$ & 0.66 \\
allure & 0.28 & $-0.01$ & $\mathbf{0.55}$ & $\mathbf{0.56}$ & 0.69 \\
analytic & $-0.11$ & $-0.12$ & 0.02 & $\mathbf{-0.48}$ & 0.26 \\
\midrule
Variance explained & 28\% & 19\% & 17\% & 7\% & \\
\bottomrule
\end{tabular}
\end{table}

\clearpage

\subsection{Features That Load on Each Factor}
\label{app:traits_loading}

\begin{table}[h]
\centering
\caption{Features load on each factor ($|\lambda| \ge 0.40$ in the varimax solution), split by the sign of the loading and ordered by its size. The two families are analyzed separately, so their factors are numbered independently; a feature may load on two factors of its family.}
\label{tab:factor_members}
\resizebox{1.0\linewidth}{!}{
\begin{tabular}{c llllll}
\toprule
& \multicolumn{2}{c}{Functional} & \multicolumn{4}{c}{Linguistic} \\
\cmidrule(lr){2-3} \cmidrule(lr){4-7}
& F1 & F2 & F1 & F2 & F3 & F4 \\
\midrule
\multirow{20}{*}{\rotatebox{90}{Positive}} & $\mathrm{trans}_{\mathrm{Eval}\to\mathrm{Plan}}$ & $\mathrm{prop}_{\mathrm{Reas}}$ & function words & clout & curiosity & negations \\
& $\mathrm{trans}_{\mathrm{Refl}\to\mathrm{Plan}}$ & $\mathrm{late}_{\mathrm{Reas}}$ & linguistic dim. & social ref. & verbs & all-or-none \\
& $\mathrm{trans}_{\mathrm{Plan}\to\mathrm{Plan}}$ & $\mathrm{mid}_{\mathrm{Reas}}$ & impersonal pron. & social proc. & personal pron. & allure \\
& $\mathrm{trans\_entropy}$ & $\mathrm{trans}_{\mathrm{Exec}\to\mathrm{Reas}}$ & present focus & polite & attention &  \\
& $\mathrm{entropy}$ & $\mathrm{early}_{\mathrm{Reas}}$ & determiners & affiliation & I &  \\
& $\mathrm{trans}_{\mathrm{Deci}\to\mathrm{Plan}}$ & $\mathrm{trans}_{\mathrm{Plan}\to\mathrm{Reas}}$ & auxiliary verbs & prosocial & total pron. &  \\
& $\mathrm{trans}_{\mathrm{Refl}\to\mathrm{Exec}}$ & $\mathrm{trans}_{\mathrm{Reas}\to\mathrm{Reas}}$ & articles & we & you &  \\
& $\mathrm{trans}_{\mathrm{Reas}\to\mathrm{Plan}}$ & $\mathrm{trans}_{\mathrm{Eval}\to\mathrm{Reas}}$ & total pron. & you & positive emotion &  \\
& $\mathrm{grps\_per\_step}$ & $\mathrm{trans}_{\mathrm{Exec}\to\mathrm{Plan}}$ & adverbs & drives (total) & emotional tone &  \\
& $\mathrm{prop}_{\mathrm{Plan}}$ & $\mathrm{grps\_per\_step}$ & prepositions & big words & insight &  \\
& $\mathrm{mid}_{\mathrm{Plan}}$ & $\mathrm{trans}_{\mathrm{Reas}\to\mathrm{Plan}}$ & achievement & attention & allure &  \\
& $\mathrm{late}_{\mathrm{Plan}}$ & $\mathrm{trans}_{\mathrm{Refl}\to\mathrm{Reas}}$ & discrepancy & future focus & work &  \\
& $\mathrm{early}_{\mathrm{Plan}}$ & $\mathrm{prop}_{\mathrm{Plan}}$ & personal pron. &  & lifestyle (total) &  \\
& $\mathrm{trans}_{\mathrm{Refl}\to\mathrm{Reas}}$ & $\mathrm{mid}_{\mathrm{Plan}}$ & we &  & need &  \\
& $\mathrm{trans}_{\mathrm{Exec}\to\mathrm{Plan}}$ & $\mathrm{early}_{\mathrm{Plan}}$ & differentiation &  & positive tone &  \\
& $\mathrm{density}$ & $\mathrm{late}_{\mathrm{Plan}}$ & adjectives &  & drives (total) &  \\
& $\mathrm{prop}_{\mathrm{Refl}}$ & $\mathrm{trans}_{\mathrm{Plan}\to\mathrm{Plan}}$ & verbs &  & future focus &  \\
& $\mathrm{trans}_{\mathrm{Eval}\to\mathrm{Reas}}$ & $\mathrm{density}$ & affiliation &  &  &  \\
& $\mathrm{trans}_{\mathrm{Reas}\to\mathrm{Reas}}$ &  &  &  &  &  \\
& $\mathrm{trans}_{\mathrm{Plan}\to\mathrm{Reas}}$ &  &  &  &  &  \\
\cmidrule(lr){1-7}
\multirow{6}{*}{\rotatebox{90}{Negative}} & $\mathrm{mid}_{\mathrm{Retr}}$ & $\mathrm{prop}_{\mathrm{Grnd}}$ & power & I & conjunctions & analytic \\
& $\mathrm{prop}_{\mathrm{Retr}}$ & $\mathrm{late}_{\mathrm{Exec}}$ & numbers & certitude & quantities & adjectives \\
& $\mathrm{late}_{\mathrm{Exec}}$ & $\mathrm{mid}_{\mathrm{Exec}}$ & big words & adverbs & differentiation & positive tone \\
& $\mathrm{mid}_{\mathrm{Exec}}$ &  & lifestyle (total) & quantities & adjectives &  \\
&  &  & work & adjectives &  &  \\
&  &  & drives (total) &  &  &  \\
\bottomrule
\end{tabular}
}
\end{table}

\clearpage

\subsection{Leave-One-Out Cross-Validation on the Factors}
\label{app:holdout_task_stability}

\paragraph{Setting.}
We ask whether the traits and factors reported in \S\ref{sec:results} depend on any single task.
For each of the 12 tasks we remove all trajectories of one task from both feature tables (e.g., 180 trajectories for \textit{gui}, 5{,}332 for \textit{swe}) and rerun the entire pipeline on the remaining eleven with unchanged settings: the harness filter, the context adjustment (the mixed model is refitted, now without the held-out task level), the three-layer tests with the same thresholds ($\rho_{\mathrm{stab}} \ge 0.7$, $\rho_{\mathrm{con}} \ge 0.3$, $\mathrm{ICC} \ge 0.1$), and factor analysis using the same configuration (Horn's parallel analysis for the number of factors, iterated PAF, varimax rotation, ridge regression scores).
Nothing is copied from the reference run: each fold selects its own traits and its own number of factors.
The fold is then compared with the reference run on two levels, the trait set and the factor structure (Table~\ref{tab:holdout-structure}).

\paragraph{Trait-level feature Jaccard.}
Let $A$ be the set of features that pass the three tests in the reference run ({\numfeatureF} functional, {\numfeatureL} linguistic) and $B$ the set that passes in the fold. The Jaccard index,
\[
J(A,B)=\frac{|A\cap B|}{|A\cup B|},
\]
is the fraction of features, among all that are a trait in at least one of the two runs, that are a trait in both.
It equals one when the two runs select exactly the same traits and zero when they share none; a feature that drops out of the trait set and one that newly enters it lower the index by the same amount.
Folds keep 22--36 functional and 43--53 linguistic traits, and the mean Jaccard is 0.87 for the functional and 0.94 for the linguistic family.
The lowest values come from holding out \textit{science} (0.71/0.77): the fold loses nine functional traits, mostly transition and stage features (e.g., Evaluate$\to$Planning, late-run Planning), and adds none, because science trajectories contribute many of the task pairs on which the cross-task consistency test is computed.
Across folds, 18 of the 31 functional and 38 of the 48 linguistic traits are selected in all 12 folds, and 25 and 45 in at least 10.

\paragraph{Tucker congruence $\phi$.}
A factor is described by its vector of loadings, one entry per trait.
To compare a factor of the reference run, $\boldsymbol{a}$, with a factor of the fold, $\boldsymbol{b}$, we restrict both to the traits kept by both runs and compute the Tucker congruence coefficient:
\[
\phi(\boldsymbol{a},\boldsymbol{b})=\frac{\sum_i a_i b_i}{\sqrt{\sum_i a_i^2\,\sum_i b_i^2}},
\]
the cosine of the angle between the two loading vectors \citep{lorenzo2006tucker}.
It is one when the fold's factor loads on the same traits in the same proportions as the reference run's (the loadings may differ by a common scale), zero when the two loading patterns are unrelated, and it ignores the sign of a factor, which is arbitrary in factor analysis (we report $|\phi|$).
Conventionally $\phi \ge 0.95$ is taken to mean the two factors are the same and $0.85$--$0.94$ that they are similar.
Because a fold may retain a different number of factors than the reference run, we first pair the factors of the two runs one-to-one so as to maximize the total congruence (Hungarian assignment).
As shown in Table~\ref{tab:holdout-structure}, two functional factors reappear in all 12 folds (mean $\phi = 0.99$, minimum $0.97$), and every fold retains exactly two of them.
Similarly, the linguistic family retains four factors in 6 of 12 folds (3--6 otherwise); its first two factors are recovered with mean $\phi = 0.97$, the third with $0.93$, and the fourth (negation, 7\% of variance in the reference run) with $0.91$, the only value below $0.85$ being the \textit{swe} fold ($\phi = 0.61$), where the fold splits into six factors.

\begin{table}[h]
\centering
\caption{Leave-one-out stability of the pipeline. For every fold the held-out task type is removed and the whole pipeline is rerun on the remaining eleven; the result is compared with the reference run. $k_F$, $k_L$: retained factors (reference run: 2 and 4). Jac.: Jaccard overlap of the fold's trait set with the reference run's ({\numfeatureF} functional, {\numfeatureL} linguistic). Per factor: Tucker congruence $\phi$ of the varimax loadings on the traits both runs kept, factors matched by Hungarian assignment; -- marks a reference-run factor without a match. Last row: mean over folds (Linguistic F4 over the 11 folds in which it is matched).}
\label{tab:holdout-structure}
\begin{tabular}{l cc cc cc cccc}
\toprule
& & & \multicolumn{2}{c}{Jac.} & \multicolumn{2}{c}{Functional $\phi$} & \multicolumn{4}{c}{Linguistic $\phi$} \\
\cmidrule(lr){4-5} \cmidrule(lr){6-7} \cmidrule(lr){8-11}
Held-out task & $k_F$ & $k_L$ & F & L & F1 & F2 & F1 & F2 & F3 & F4 \\
\midrule
swe            & 2 & 6 & .87 & .92 & .99 & .98 & .89 & .93 & .92 & .61 \\
terminal       & 2 & 4 & .81 & .86 & 1.00 & 1.00 & .99 & .99 & .98 & .99 \\
tool-api       & 2 & 5 & .76 & .91 & .97 & .97 & .95 & .93 & .86 & .88 \\
web            & 2 & 4 & .94 & 1.00 & 1.00 & 1.00 & 1.00 & 1.00 & 1.00 & 1.00 \\
gui            & 2 & 5 & 1.00 & 1.00 & 1.00 & 1.00 & .96 & .97 & .91 & .94 \\
office         & 2 & 4 & .94 & 1.00 & 1.00 & .99 & .95 & .99 & .91 & .99 \\
research-qa    & 2 & 5 & .88 & 1.00 & 1.00 & 1.00 & .97 & .98 & .92 & .93 \\
science        & 2 & 4 & .71 & .77 & .98 & .99 & .98 & .94 & .89 & .76 \\
security       & 2 & 4 & .84 & .96 & 1.00 & 1.00 & 1.00 & 1.00 & 1.00 & .99 \\
ml-engineering & 2 & 4 & .84 & .98 & 1.00 & 1.00 & 1.00 & 1.00 & 1.00 & .99 \\
dialogue       & 2 & 3 & .91 & .94 & 1.00 & 1.00 & .97 & .97 & .94 & -- \\
other          & 2 & 5 & .94 & .98 & 1.00 & 1.00 & .94 & .97 & .89 & .93 \\
\midrule
Mean           & 2.0 & 4.4 & .87 & .94 & .99 & .99 & .97 & .97 & .93 & .91 \\
\bottomrule
\end{tabular}
\end{table}

\clearpage

\subsection{Model Traits}

\definecolor{pos}{HTML}{EB6834}
\definecolor{neg}{HTML}{2A78D6}
\begin{table}[h]
\centering
\caption{Factor scores of all models (standardized across models to zero mean and unit variance), grouped by model family. Cells are shaded by the score, orange for positive and blue for negative, with full saturation at $|z| = 3$.}
\label{tab:factor_scores_all}
\resizebox{1.0\linewidth}{!}{
\begin{tabular}{p{6cm} rrrrrr}
\toprule
Model & Fun\_F1 & Fun\_F2 & Lin\_F1 & Lin\_F2 & Lin\_F3 & Lin\_F4 \\
\midrule
alibaba/qwen2.5-72b-instruct & \cellcolor{pos!1}0.05 & \cellcolor{pos!34}1.46 & \cellcolor{pos!47}2.02 & \cellcolor{pos!20}0.87 & \cellcolor{pos!11}0.46 & \cellcolor{pos!2}0.09 \\
alibaba/qwen3-coder-480b-a35b-instruct & \cellcolor{pos!22}0.95 & \cellcolor{neg!19}-0.82 & \cellcolor{neg!10}-0.41 & \cellcolor{neg!4}-0.18 & \cellcolor{pos!30}1.28 & \cellcolor{pos!7}0.29 \\
alibaba/qwen3.5-27b & \cellcolor{pos!16}0.68 & \cellcolor{neg!10}-0.41 & \cellcolor{pos!10}0.41 & \cellcolor{neg!27}-1.18 & \cellcolor{pos!5}0.23 & \cellcolor{neg!18}-0.75 \\
alibaba/qwen3.5-35b-a3b & \cellcolor{pos!37}1.57 & \cellcolor{neg!10}-0.42 & \cellcolor{pos!7}0.31 & \cellcolor{neg!22}-0.94 & \cellcolor{neg!1}-0.06 & \cellcolor{neg!28}-1.21 \\
alibaba/qwen3.5-397b-a17b & \cellcolor{pos!26}1.10 & \cellcolor{neg!11}-0.49 & \cellcolor{pos!9}0.38 & \cellcolor{neg!23}-1.00 & \cellcolor{neg!3}-0.13 & \cellcolor{neg!17}-0.74 \\
alibaba/qwen3.6-27b & \cellcolor{pos!12}0.53 & \cellcolor{neg!8}-0.32 & \cellcolor{pos!16}0.68 & \cellcolor{neg!36}-1.54 & \cellcolor{neg!3}-0.14 & \cellcolor{neg!6}-0.28 \\
alibaba/qwen3.6-35b-a3b & \cellcolor{pos!18}0.77 & \cellcolor{neg!2}-0.09 & \cellcolor{pos!13}0.57 & \cellcolor{neg!27}-1.16 & \cellcolor{neg!20}-0.86 & \cellcolor{neg!10}-0.42 \\
alibaba/qwen3.6-plus & \cellcolor{pos!18}0.76 & \cellcolor{neg!2}-0.08 & \cellcolor{pos!12}0.52 & \cellcolor{neg!25}-1.06 & \cellcolor{neg!17}-0.75 & \cellcolor{neg!10}-0.44 \\
alibaba/qwen3.7-plus & \cellcolor{pos!25}1.08 & \cellcolor{neg!4}-0.18 & \cellcolor{pos!5}0.23 & \cellcolor{neg!35}-1.48 & \cellcolor{neg!12}-0.52 & \cellcolor{pos!1}0.04 \\
\midrule
anthropic/claude-3-5-sonnet-20240620 & \cellcolor{pos!23}0.97 & \cellcolor{neg!10}-0.45 & \cellcolor{pos!25}1.06 & \cellcolor{pos!70}4.33 & \cellcolor{neg!6}-0.24 & \cellcolor{neg!31}-1.34 \\
anthropic/claude-3-5-sonnet-20241022 & \cellcolor{neg!22}-0.96 & \cellcolor{pos!22}0.95 & \cellcolor{pos!9}0.38 & \cellcolor{pos!39}1.67 & \cellcolor{pos!8}0.33 & \cellcolor{neg!5}-0.23 \\
anthropic/claude-3-7-sonnet-20250219 & \cellcolor{neg!3}-0.13 & \cellcolor{pos!2}0.07 & \cellcolor{pos!9}0.37 & \cellcolor{pos!20}0.84 & \cellcolor{pos!25}1.09 & \cellcolor{neg!12}-0.50 \\
anthropic/claude-haiku-4-5-20251001 & \cellcolor{pos!4}0.15 & \cellcolor{neg!24}-1.03 & \cellcolor{neg!10}-0.43 & \cellcolor{neg!3}-0.12 & \cellcolor{pos!27}1.17 & \cellcolor{neg!14}-0.60 \\
anthropic/claude-opus-4-1-20250805 & \cellcolor{neg!2}-0.10 & \cellcolor{neg!14}-0.60 & \cellcolor{neg!15}-0.65 & \cellcolor{pos!3}0.14 & \cellcolor{pos!29}1.22 & \cellcolor{neg!14}-0.62 \\
anthropic/claude-opus-4-20250514 & \cellcolor{neg!10}-0.43 & \cellcolor{pos!9}0.39 & \cellcolor{pos!8}0.34 & \cellcolor{pos!1}0.04 & \cellcolor{pos!33}1.39 & \cellcolor{neg!8}-0.32 \\
anthropic/claude-opus-4-5 & \cellcolor{neg!16}-0.69 & \cellcolor{neg!4}-0.17 & \cellcolor{pos!1}0.03 & \cellcolor{neg!11}-0.48 & \cellcolor{neg!3}-0.15 & \cellcolor{pos!56}2.38 \\
anthropic/claude-opus-4-5-20251101 & \cellcolor{neg!1}-0.03 & \cellcolor{neg!14}-0.61 & \cellcolor{neg!13}-0.57 & \cellcolor{neg!4}-0.16 & \cellcolor{pos!13}0.55 & \cellcolor{neg!15}-0.66 \\
anthropic/claude-opus-4-6 & \cellcolor{neg!8}-0.34 & \cellcolor{pos!6}0.27 & \cellcolor{neg!27}-1.17 & \cellcolor{neg!16}-0.68 & \cellcolor{neg!2}-0.10 & \cellcolor{neg!2}-0.11 \\
anthropic/claude-opus-4-7 & \cellcolor{pos!12}0.53 & \cellcolor{neg!16}-0.68 & \cellcolor{pos!15}0.63 & \cellcolor{neg!29}-1.23 & \cellcolor{neg!23}-1.00 & \cellcolor{neg!31}-1.34 \\
anthropic/claude-opus-4-8 & \cellcolor{pos!39}1.68 & \cellcolor{neg!11}-0.45 & \cellcolor{pos!9}0.39 & \cellcolor{neg!18}-0.78 & \cellcolor{neg!34}-1.46 & \cellcolor{neg!11}-0.47 \\
anthropic/claude-sonnet-4-20250514 & \cellcolor{pos!3}0.13 & \cellcolor{pos!6}0.25 & \cellcolor{pos!5}0.24 & \cellcolor{neg!4}-0.18 & \cellcolor{pos!35}1.50 & \cellcolor{neg!11}-0.47 \\
anthropic/claude-sonnet-4-5-20250929 & \cellcolor{pos!4}0.19 & \cellcolor{neg!17}-0.74 & \cellcolor{neg!13}-0.57 & \cellcolor{neg!0}-0.01 & \cellcolor{pos!28}1.18 & \cellcolor{neg!12}-0.51 \\
anthropic/claude-sonnet-4-6 & \cellcolor{pos!46}1.98 & \cellcolor{neg!25}-1.07 & \cellcolor{pos!11}0.46 & \cellcolor{neg!30}-1.30 & \cellcolor{neg!18}-0.79 & \cellcolor{neg!23}-0.96 \\
\midrule
deepseek/deepseek-r1 & \cellcolor{neg!27}-1.15 & \cellcolor{pos!13}0.56 & \cellcolor{pos!7}0.30 & \cellcolor{pos!11}0.48 & \cellcolor{neg!5}-0.21 & \cellcolor{pos!20}0.87 \\
deepseek/deepseek-v2.5 & \cellcolor{neg!2}-0.09 & \cellcolor{pos!40}1.71 & \cellcolor{pos!53}2.25 & \cellcolor{pos!29}1.26 & \cellcolor{pos!12}0.50 & \cellcolor{neg!1}-0.02 \\
deepseek/deepseek-v3 & \cellcolor{neg!17}-0.73 & \cellcolor{neg!1}-0.04 & \cellcolor{neg!8}-0.34 & \cellcolor{pos!10}0.44 & \cellcolor{pos!14}0.61 & \cellcolor{neg!11}-0.49 \\
deepseek/deepseek-v3-0324 & \cellcolor{neg!28}-1.19 & \cellcolor{neg!13}-0.54 & \cellcolor{neg!17}-0.74 & \cellcolor{pos!1}0.06 & \cellcolor{pos!17}0.72 & \cellcolor{pos!9}0.37 \\
deepseek/deepseek-v3.1 & \cellcolor{neg!12}-0.53 & \cellcolor{pos!1}0.04 & \cellcolor{neg!19}-0.83 & \cellcolor{neg!4}-0.16 & \cellcolor{pos!20}0.85 & \cellcolor{neg!9}-0.37 \\
deepseek/deepseek-v3.2 & \cellcolor{pos!4}0.16 & \cellcolor{neg!3}-0.14 & \cellcolor{pos!10}0.41 & \cellcolor{pos!10}0.44 & \cellcolor{neg!12}-0.51 & \cellcolor{pos!70}3.49 \\
deepseek/deepseek-v3.2-exp & \cellcolor{pos!21}0.92 & \cellcolor{neg!29}-1.26 & \cellcolor{neg!3}-0.13 & \cellcolor{neg!11}-0.46 & \cellcolor{pos!45}1.92 & \cellcolor{pos!19}0.83 \\
deepseek/deepseek-v4-flash & \cellcolor{pos!15}0.65 & \cellcolor{pos!5}0.23 & \cellcolor{pos!15}0.63 & \cellcolor{neg!29}-1.23 & \cellcolor{neg!14}-0.60 & \cellcolor{neg!7}-0.30 \\
deepseek/deepseek-v4-pro & \cellcolor{pos!30}1.29 & \cellcolor{pos!9}0.38 & \cellcolor{pos!15}0.63 & \cellcolor{neg!29}-1.23 & \cellcolor{neg!25}-1.05 & \cellcolor{neg!7}-0.29 \\
\midrule
google/gemini-2.0-flash & \cellcolor{neg!4}-0.16 & \cellcolor{neg!11}-0.46 & \cellcolor{pos!4}0.17 & \cellcolor{pos!1}0.06 & \cellcolor{pos!24}1.04 & \cellcolor{pos!1}0.03 \\
google/gemini-2.5-flash & \cellcolor{neg!23}-0.98 & \cellcolor{neg!36}-1.53 & \cellcolor{neg!9}-0.37 & \cellcolor{pos!5}0.22 & \cellcolor{pos!17}0.72 & \cellcolor{pos!5}0.20 \\
google/gemini-2.5-pro & \cellcolor{pos!1}0.06 & \cellcolor{neg!33}-1.41 & \cellcolor{neg!5}-0.20 & \cellcolor{neg!1}-0.03 & \cellcolor{pos!20}0.87 & \cellcolor{neg!2}-0.07 \\
google/gemini-3-flash-preview & \cellcolor{neg!4}-0.19 & \cellcolor{neg!10}-0.45 & \cellcolor{pos!2}0.10 & \cellcolor{neg!11}-0.49 & \cellcolor{pos!14}0.61 & \cellcolor{pos!22}0.94 \\
google/gemini-3-pro-preview & \cellcolor{neg!28}-1.20 & \cellcolor{neg!16}-0.71 & \cellcolor{neg!8}-0.33 & \cellcolor{neg!6}-0.27 & \cellcolor{neg!9}-0.41 & \cellcolor{pos!70}4.18 \\
google/gemini-3.1-pro-preview & \cellcolor{neg!11}-0.49 & \cellcolor{pos!36}1.55 & \cellcolor{neg!5}-0.20 & \cellcolor{neg!10}-0.41 & \cellcolor{pos!15}0.64 & \cellcolor{neg!13}-0.55 \\
\midrule
meta/llama-3.1-405b-instruct & \cellcolor{pos!4}0.17 & \cellcolor{pos!46}1.96 & \cellcolor{pos!67}2.87 & \cellcolor{neg!0}-0.01 & \cellcolor{pos!32}1.37 & \cellcolor{pos!17}0.71 \\
meta/llama-3.1-70b-instruct & \cellcolor{neg!21}-0.91 & \cellcolor{pos!70}3.13 & \cellcolor{pos!62}2.64 & \cellcolor{neg!8}-0.33 & \cellcolor{pos!31}1.34 & \cellcolor{pos!3}0.14 \\
meta/llama-3.3-70b-instruct & \cellcolor{pos!16}0.67 & \cellcolor{pos!39}1.68 & \cellcolor{pos!60}2.57 & \cellcolor{neg!15}-0.66 & \cellcolor{pos!32}1.37 & \cellcolor{pos!17}0.73 \\
\midrule
minimax/minimax-m2 & \cellcolor{pos!21}0.90 & \cellcolor{neg!4}-0.18 & \cellcolor{neg!26}-1.10 & \cellcolor{neg!14}-0.59 & \cellcolor{pos!36}1.55 & \cellcolor{neg!4}-0.17 \\
minimax/minimax-m2.1 & \cellcolor{neg!3}-0.12 & \cellcolor{neg!13}-0.57 & \cellcolor{neg!33}-1.42 & \cellcolor{pos!7}0.29 & \cellcolor{pos!39}1.66 & \cellcolor{neg!35}-1.50 \\
minimax/minimax-m2.5 & \cellcolor{pos!7}0.31 & \cellcolor{pos!31}1.31 & \cellcolor{neg!16}-0.69 & \cellcolor{neg!13}-0.58 & \cellcolor{pos!25}1.08 & \cellcolor{pos!8}0.35 \\
minimax/minimax-m2.7 & \cellcolor{pos!24}1.02 & \cellcolor{pos!60}2.57 & \cellcolor{pos!19}0.80 & \cellcolor{neg!30}-1.28 & \cellcolor{neg!19}-0.82 & \cellcolor{neg!2}-0.10 \\
minimax/minimax-m3 & \cellcolor{pos!26}1.13 & \cellcolor{pos!8}0.35 & \cellcolor{pos!18}0.79 & \cellcolor{neg!26}-1.13 & \cellcolor{neg!28}-1.22 & \cellcolor{neg!7}-0.31 \\
\midrule
\end{tabular}
}
\end{table}
\begin{table}[t]
\centering
\resizebox{1.0\linewidth}{!}{
\begin{tabular}{p{6cm} rrrrrr}
\multicolumn{7}{l}{\textit{\small Table Continued}} \\
\toprule
Model & Fun\_F1 & Fun\_F2 & Lin\_F1 & Lin\_F2 & Lin\_F3 & Lin\_F4 \\
\midrule
moonshot/kimi-k2-0905-preview & \cellcolor{pos!23}0.98 & \cellcolor{neg!25}-1.08 & \cellcolor{neg!13}-0.55 & \cellcolor{neg!8}-0.36 & \cellcolor{pos!31}1.32 & \cellcolor{neg!1}-0.06 \\
moonshot/kimi-k2-thinking & \cellcolor{pos!22}0.95 & \cellcolor{neg!18}-0.79 & \cellcolor{neg!19}-0.80 & \cellcolor{neg!11}-0.46 & \cellcolor{pos!22}0.95 & \cellcolor{pos!3}0.13 \\
moonshot/kimi-k2.5 & \cellcolor{pos!22}0.94 & \cellcolor{pos!11}0.46 & \cellcolor{neg!6}-0.27 & \cellcolor{neg!14}-0.62 & \cellcolor{neg!9}-0.37 & \cellcolor{pos!38}1.63 \\
moonshot/kimi-k2.6 & \cellcolor{pos!7}0.32 & \cellcolor{pos!12}0.50 & \cellcolor{pos!10}0.44 & \cellcolor{neg!5}-0.19 & \cellcolor{neg!37}-1.58 & \cellcolor{neg!17}-0.73 \\
moonshot/kimi-k2.7-code & \cellcolor{pos!11}0.46 & \cellcolor{pos!70}3.22 & \cellcolor{pos!18}0.77 & \cellcolor{neg!11}-0.45 & \cellcolor{neg!33}-1.43 & \cellcolor{pos!20}0.88 \\
moonshot/kimi-k3 & \cellcolor{pos!40}1.72 & \cellcolor{neg!7}-0.29 & \cellcolor{neg!11}-0.46 & \cellcolor{neg!13}-0.57 & \cellcolor{neg!46}-1.97 & \cellcolor{neg!23}-1.00 \\
\midrule
openai/gpt-4.1-2025-04-14 & \cellcolor{neg!26}-1.10 & \cellcolor{neg!1}-0.05 & \cellcolor{neg!4}-0.16 & \cellcolor{pos!21}0.89 & \cellcolor{pos!1}0.03 & \cellcolor{neg!9}-0.40 \\
openai/gpt-4.1-mini-2025-04-14 & \cellcolor{neg!28}-1.18 & \cellcolor{neg!7}-0.31 & \cellcolor{neg!16}-0.69 & \cellcolor{pos!25}1.07 & \cellcolor{neg!10}-0.44 & \cellcolor{neg!8}-0.34 \\
openai/gpt-4o-2024-05-13 & \cellcolor{pos!17}0.73 & \cellcolor{pos!51}2.20 & \cellcolor{pos!70}3.17 & \cellcolor{pos!70}3.23 & \cellcolor{neg!1}-0.03 & \cellcolor{neg!14}-0.60 \\
openai/gpt-5-2025-08-07 & \cellcolor{neg!29}-1.25 & \cellcolor{neg!22}-0.96 & \cellcolor{neg!26}-1.12 & \cellcolor{pos!14}0.58 & \cellcolor{neg!13}-0.56 & \cellcolor{neg!2}-0.10 \\
openai/gpt-5-codex & \cellcolor{neg!27}-1.14 & \cellcolor{pos!7}0.31 & \cellcolor{neg!43}-1.86 & \cellcolor{pos!18}0.75 & \cellcolor{pos!8}0.36 & \cellcolor{neg!2}-0.09 \\
openai/gpt-5-mini-2025-08-07 & \cellcolor{neg!34}-1.46 & \cellcolor{neg!34}-1.44 & \cellcolor{neg!29}-1.22 & \cellcolor{pos!14}0.62 & \cellcolor{neg!9}-0.37 & \cellcolor{neg!2}-0.10 \\
openai/gpt-5-nano-2025-08-07 & \cellcolor{neg!27}-1.15 & \cellcolor{neg!35}-1.51 & \cellcolor{neg!22}-0.93 & \cellcolor{pos!20}0.87 & \cellcolor{neg!23}-0.97 & \cellcolor{neg!11}-0.46 \\
openai/gpt-5.1 & \cellcolor{neg!62}-2.65 & \cellcolor{neg!27}-1.16 & \cellcolor{neg!19}-0.83 & \cellcolor{pos!22}0.95 & \cellcolor{neg!29}-1.24 & \cellcolor{pos!16}0.67 \\
openai/gpt-5.1-codex & \cellcolor{neg!19}-0.81 & \cellcolor{pos!24}1.05 & \cellcolor{neg!35}-1.50 & \cellcolor{pos!17}0.72 & \cellcolor{pos!0}0.00 & \cellcolor{pos!15}0.63 \\
openai/gpt-5.2 & \cellcolor{neg!38}-1.62 & \cellcolor{neg!15}-0.64 & \cellcolor{neg!19}-0.82 & \cellcolor{pos!15}0.65 & \cellcolor{neg!30}-1.28 & \cellcolor{pos!36}1.56 \\
openai/gpt-5.3-codex & \cellcolor{neg!21}-0.89 & \cellcolor{pos!4}0.17 & \cellcolor{neg!16}-0.69 & \cellcolor{pos!18}0.75 & \cellcolor{neg!15}-0.65 & \cellcolor{pos!11}0.48 \\
openai/gpt-5.5 & \cellcolor{pos!17}0.71 & \cellcolor{pos!2}0.08 & \cellcolor{pos!17}0.73 & \cellcolor{neg!0}-0.01 & \cellcolor{neg!43}-1.85 & \cellcolor{neg!14}-0.60 \\
openai/gpt-5.6-sol & \cellcolor{pos!20}0.86 & \cellcolor{neg!27}-1.16 & \cellcolor{neg!32}-1.39 & \cellcolor{pos!22}0.93 & \cellcolor{neg!52}-2.22 & \cellcolor{neg!43}-1.85 \\
openai/gpt-oss-120b & \cellcolor{neg!30}-1.28 & \cellcolor{neg!7}-0.31 & \cellcolor{neg!17}-0.71 & \cellcolor{pos!31}1.34 & \cellcolor{neg!17}-0.73 & \cellcolor{pos!9}0.41 \\
openai/gpt-oss-20b & \cellcolor{neg!40}-1.71 & \cellcolor{pos!14}0.62 & \cellcolor{neg!5}-0.21 & \cellcolor{pos!37}1.57 & \cellcolor{neg!33}-1.43 & \cellcolor{pos!66}2.82 \\
openai/o3-2025-04-16 & \cellcolor{neg!39}-1.65 & \cellcolor{neg!6}-0.27 & \cellcolor{neg!15}-0.64 & \cellcolor{pos!13}0.55 & \cellcolor{neg!2}-0.10 & \cellcolor{neg!8}-0.36 \\
openai/o3-mini & \cellcolor{neg!31}-1.32 & \cellcolor{pos!0}0.00 & \cellcolor{pos!1}0.03 & \cellcolor{pos!28}1.19 & \cellcolor{neg!1}-0.02 & \cellcolor{neg!16}-0.69 \\
openai/o4-mini-2025-04-16 & \cellcolor{neg!41}-1.75 & \cellcolor{neg!0}-0.02 & \cellcolor{neg!9}-0.40 & \cellcolor{pos!15}0.66 & \cellcolor{pos!7}0.30 & \cellcolor{neg!11}-0.48 \\
\midrule
xai/grok-4-0709 & \cellcolor{neg!21}-0.91 & \cellcolor{neg!19}-0.80 & \cellcolor{neg!23}-0.99 & \cellcolor{pos!9}0.39 & \cellcolor{neg!2}-0.08 & \cellcolor{pos!15}0.66 \\
xai/grok-code-fast-1 & \cellcolor{neg!16}-0.69 & \cellcolor{neg!11}-0.46 & \cellcolor{neg!22}-0.95 & \cellcolor{pos!6}0.26 & \cellcolor{neg!4}-0.19 & \cellcolor{pos!17}0.73 \\
\midrule
xiaomi/mimo-v2.5-pro & \cellcolor{pos!23}0.97 & \cellcolor{neg!7}-0.29 & \cellcolor{pos!10}0.42 & \cellcolor{neg!23}-0.98 & \cellcolor{neg!20}-0.85 & \cellcolor{neg!16}-0.70 \\
\midrule
zhipu/glm-4.6 & \cellcolor{pos!26}1.11 & \cellcolor{neg!17}-0.71 & \cellcolor{neg!13}-0.56 & \cellcolor{neg!11}-0.48 & \cellcolor{pos!37}1.58 & \cellcolor{neg!1}-0.04 \\
zhipu/glm-4.7 & \cellcolor{pos!5}0.22 & \cellcolor{pos!14}0.60 & \cellcolor{neg!20}-0.84 & \cellcolor{neg!10}-0.44 & \cellcolor{neg!3}-0.13 & \cellcolor{neg!11}-0.48 \\
zhipu/glm-5 & \cellcolor{pos!18}0.78 & \cellcolor{pos!1}0.04 & \cellcolor{neg!22}-0.94 & \cellcolor{neg!8}-0.35 & \cellcolor{pos!12}0.50 & \cellcolor{pos!7}0.31 \\
zhipu/glm-5.1 & \cellcolor{pos!17}0.74 & \cellcolor{neg!1}-0.02 & \cellcolor{pos!10}0.43 & \cellcolor{neg!19}-0.83 & \cellcolor{neg!33}-1.40 & \cellcolor{neg!20}-0.84 \\
zhipu/glm-5.2 & \cellcolor{pos!26}1.10 & \cellcolor{pos!2}0.09 & \cellcolor{pos!13}0.54 & \cellcolor{neg!24}-1.01 & \cellcolor{neg!32}-1.38 & \cellcolor{neg!11}-0.47 \\
\midrule
\end{tabular}
}
\end{table}

\end{document}